\documentclass[10pt,journal,compsoc]{IEEEtran}

\ifCLASSOPTIONcompsoc
  \usepackage[nocompress]{cite}
\else
  \usepackage{cite}
\fi

\usepackage{amsmath,amsfonts}
\usepackage[nice]{nicefrac}
\usepackage[caption=false,font=normalsize,labelfont=sf,textfont=sf]{subfig}
\usepackage{graphicx}
\usepackage{multirow}
\usepackage{tikz}
\usepackage{amssymb}
\usepackage{hyperref}
\hypersetup{
    breaklinks=false,   
    colorlinks=false,   
    pdfusetitle=true,   
}
\begin{document}
\title{Training Image Derivatives: Increased Accuracy and Universal Robustness}
\author{Vsevolod~I.~Avrutskiy
\IEEEcompsocitemizethanks{\IEEEcompsocthanksitem V.I. Avrutskiy is an independent researcher (e-mail: vsevolod@pm.me)\protect\\
}
}


\IEEEtitleabstractindextext{%
\begin{abstract}
Derivative training is an established method that can significantly increase the accuracy of neural networks in certain low-dimensional tasks. In this paper, we extend this improvement to an illustrative image analysis problem: reconstructing the vertices of a cube from its image. By training the derivatives with respect to the cube's six degrees of freedom, we achieve a 25-fold increase in accuracy for noiseless inputs. Additionally, derivative knowledge offers a novel approach to enhancing network robustness, which has traditionally been understood in terms of two types of vulnerabilities: excessive sensitivity to minor perturbations and failure to detect significant image changes. Conventional robust training relies on output invariance, which inherently creates a trade-off between these two vulnerabilities. By leveraging derivative information we compute non-trivial output changes in response to arbitrary input perturbations. This resolves the trade-off, yielding a network that is twice as robust and five times more accurate than the best case under the invariance assumption. Unlike conventional robust training, this outcome can be further improved by simply increasing the network capacity. This approach is applicable to phase retrieval problems and other scenarios where a sufficiently smooth manifold parametrization can be obtained.\end{abstract}

}
\maketitle
\section{Introduction}
\IEEEPARstart{N}{eural} networks can solve a wide range of problems, and their use is supported by theorems \cite{hornik_approximation_1991,kuurkova_kolmogorovs_1992,barron_approximation_1994} that guarantee the existence of parameters which yield arbitrarily accurate solutions. The most prominent use cases are high-dimensional
approximations that require deep architectures \cite{lecun_deep_2015}.
For deep networks, the main tool for tuning parameters is gradient-based
optimization, which minimizes the deviation of the network from the
target function. However, the local nature of this method can prevent
it from achieving arbitrarily low error even in simple cases.
In the previous paper \cite{avrutskiy_enhancing_2020}, we demonstrated that the accuracy of gradient-based training can be significantly improved (by several orders of magnitude) when the deviations of the network derivatives from the target derivatives are minimized. This method has been shown to work well for relatively
low-dimensional applications such as solving partial differential equations 
\cite{avrutskiy_neural_2020}, computational chemistry \cite{montes-campos_differentiable_2021,carrete_deep_2023},
robotic control \cite{kim_learning_2022,wang_neural_2022}, transport
coefficients evaluation \cite{istomin_evaluation_2023}, and Bayesian
inference \cite{kwok_laplace_2023}. These improvements have not been demonstrated in high-dimensional cases like image analysis, and this paper establishes the effectiveness of the method for such applications.

In high-dimensional cases, the problem of accuracy takes a slightly different form. The data typically belong to a lower-dimensional manifold \cite{cayton_algorithms_2008,fefferman_testing_2016}, and high accuracy on this manifold can be entirely compromised by small perturbations from the full input space. Practical accuracy is, therefore, inextricably linked to resilience to perturbations. 
This resilience is typically measured by how well the system responds to adversarial examples - inputs deliberately crafted to deceive neural networks.
There are two fundamental types of adversarial examples. The first exploits
excessive sensitivity: the network's output is altered, even though
the input changes are imperceptible \cite{szegedy_intriguing_2013}.
The network gradient contains all the information about
local sensitivity \cite{lyu_unified_2015} and can be used to
design attacks and train the network to withstand them \cite{goodfellow_explaining_2014,bai_recent_2021}.
The second type exploits invariance: the network's output remains the same, despite the essential alterations to the
input \cite{jacobsen_excessive_2018,jacobsen_exploiting_2019}. The
``essentiality'' of the alterations can only be determined by an ideal target function (an oracle), making human evaluation necessary for generating such adversarial examples \cite{tramer_fundamental_2020}. This complicates both the analysis of network vulnerabilities and the development of defenses, which remains an open problem. Although the relationship between accuracy and robustness is far from obvious \cite{su_is_2018,tsipras_robustness_2018,zhang_theoretically_2019,kamath_can_2021,moayeri_explicit_2022},
recent work has shown that high accuracy is compatible with robustness
against sensitivity attacks \cite{stutz_disentangling_2019,nakkiran_adversarial_2019},
but there is an inherent trade-off with the ability to withstand invariance attacks
\cite{tramer_fundamental_2020,dobriban_provable_2023}. Several studies
consider accuracy and different types of robustness as conflicting
goals and propose Pareto optimization \cite{sun_ke_pareto_2023} and
other types of trade-offs \cite{chen_balanced_2022,yang_one_2022,rauter_effect_2023}.

This paper presents an illustrative image analysis problem in which the geometry of the data manifold is accessible. This allows for the construction of an oracle defined over the entire input space and the computation of its Taylor expansion in the vicinity of the manifold. The expansion defines a non-trivial output response to any input perturbation, enabling genuinely robust training without the inherent trade-off. It also facilitates a robustness analysis that unifies sensitivity-based and invariance-based attacks. Furthermore, we demonstrate that training derivatives can significantly improve accuracy in the non-robust formulation, extending the results obtained for low-dimensional cases \cite{avrutskiy_enhancing_2020, avrutskiy_neural_2020} to image-input neural networks. The presented technique is applicable to real-world problems where the data manifold is computationally accessible, such as phase retrieval \cite{shechtman_phase_2015, yao_autophasenn_2022}, denoising images generated through differentiable rendering \cite{kato_differentiable_2020}, or when a sufficiently smooth latent representation exists.

The structure of the paper is as follows: Section \ref{sec:Problem-formulation} formulates the problem and discusses the results of conventional non-robust training. A definition of the oracle is then provided.
Section \ref{sec:first-order} focuses on the first order, beginning with the computation of image derivatives to improve the accuracy of the non-robust problem. The first-order expansion of the oracle is then obtained, and a robustness analysis is formulated. First-order robust training is proposed, implemented and compared against Gaussian augmentation and Jacobian regularization. Section \ref{sec:second-order} follows a similar structure, focusing on the second order. The remaining sections are self-explanatory.

\section{Problem formulation\label{sec:Problem-formulation}}
\subsection{Notations}
Images are represented as vectors in Euclidean space and are denoted by
calligraphic letters. Images from the manifold are denoted by $\mathcal{I}$.
Images from the full image space, with no restrictions on pixel values,
are denoted by $\mathcal{F}$. The scalar product is denoted by a
dot $\mathcal{I}\cdot\mathcal{F}$ and the norm is $\left\Vert \mathcal{I}\right\Vert =\sqrt{\mathcal{I}\cdot\mathcal{I}}$.
For low-dimensional vectors, index notation is used $r_{i} \equiv \vec{r}$.
Whenever indices are repeated in a single term, summation is implied,
for example, in the scalar product
\[
r_{i}s_{i}\equiv\sum_{i}r_{i}s_{i}
\]
or in squared norm $r_{i}r_{i}\equiv\left\Vert \vec{r}\right\Vert ^{2}$.
Terms with non-repeating indices are tensors of higher rank, e.g.
the outer product of two 3D vectors $r_{i}s_{j}$ is a 3 by 3 matrix.
The symbol $\delta_{ji}$ is the Kronecker delta
\[
\delta_{ji}=\begin{cases}
0 & i\neq j\\
1 & i=j
\end{cases}
\]

\subsection{Problem Formulation}
\begin{figure}
\centering
\includegraphics[height=3cm]{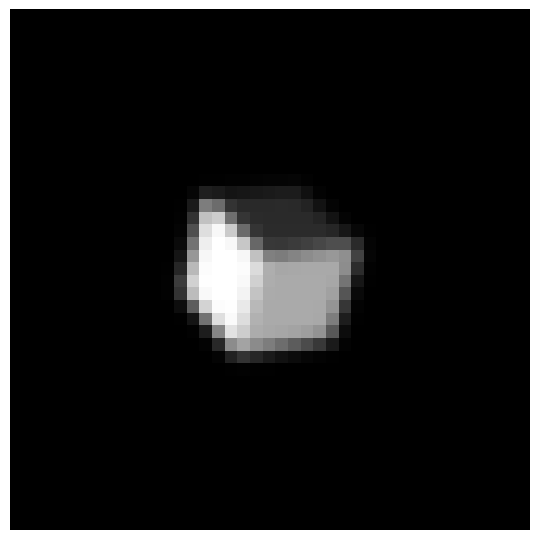}\includegraphics[height=3cm]{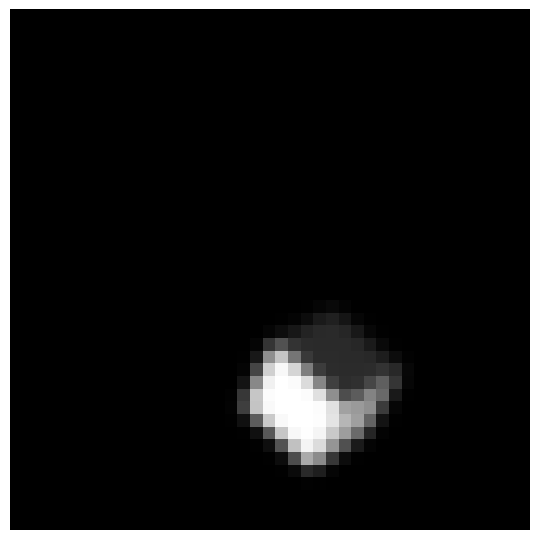}\includegraphics[height=3cm]{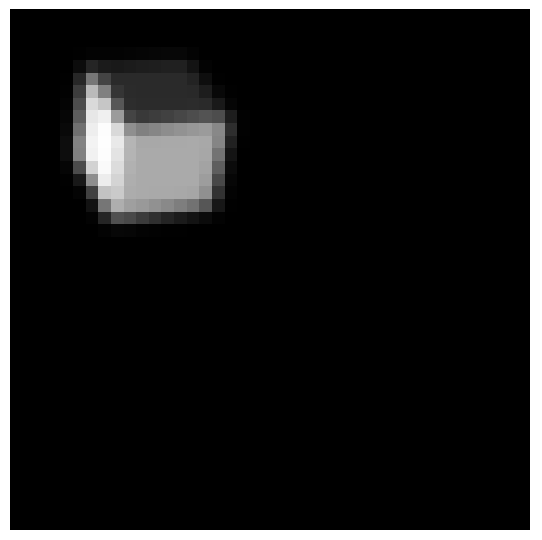}

\includegraphics[height=3cm]{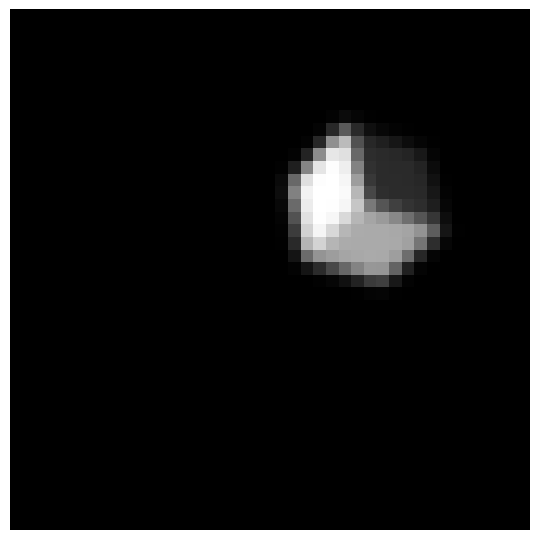}\includegraphics[height=3cm]{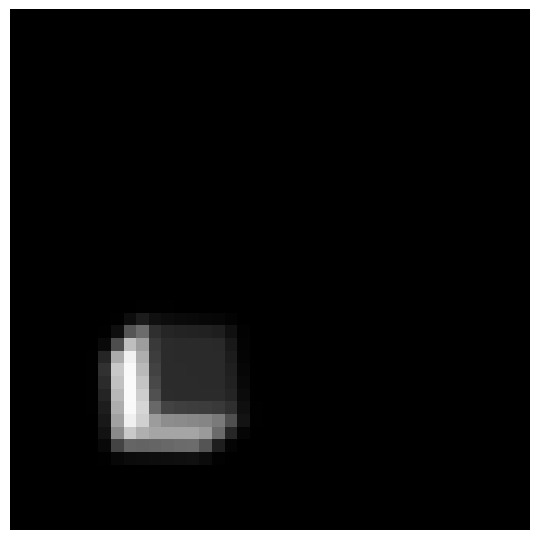}\includegraphics[height=3cm]{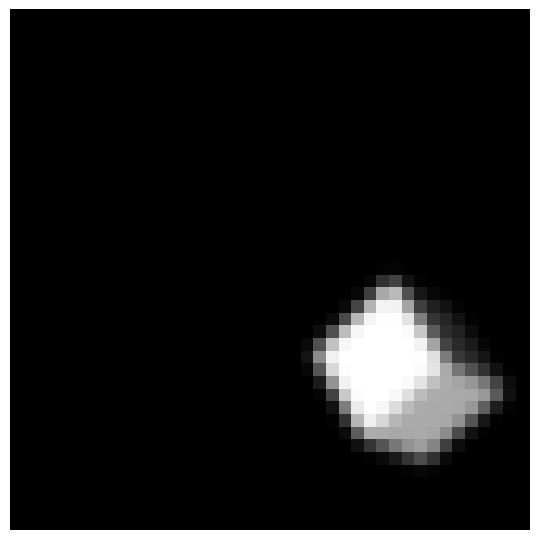}

\caption{Images of the cube, 41 by 41 pixels\label{fig:Cube-images}}
\end{figure}

Consider a cube $C$ and an algorithm that renders its image $\mathcal{I}\left(C\right)$. For simplicity, we construct a straightforward differentiable algorithm from scratch, with all details provided in Appendix \ref{Image-gen}. The resulting 41 by 41 pixel images are shown in Fig. \ref{fig:Cube-images}. Differentiable rendering \cite{kato_differentiable_2020} can be employed for more complex scenarios.

The cube $C$ has 6 degrees of freedom, and its images lie on a 6-dimensional
manifold. The goal is to determine the coordinates of the cube's vertices
\[
C=\left(\begin{array}{ccc}
x_{1} & \ldots & x_{8}\\
y_{1} & \ldots & y_{8}\\
z_{1} & \ldots & z_{8}
\end{array}\right)
\]
from its image. We aim to train a neural network $N$ with 9 outputs to determine the coordinates of the first three vertices
\[
N\left(\mathcal{I}\left(C\right)\right)=\left(x_{1},y_{1},z_{1},x_{2},y_{2},z_{2},x_{3},y_{3},z_{3}\right),
\]
and since they are ordered, the remaining ones can be easily reconstructed.
For brevity, we will use this 9-dimensional vector interchangeably with the cube $C$, so the task is
\begin{equation}
N\left(\mathcal{I}\left(C\right)\right)=C\label{eq:problem},
\end{equation}
and target components are denoted by $C_{i}$, $i\in\left[1,9\right]$. The cost function is
\begin{equation}
E_{0}=\frac{1}{n^{2}}\sum_{\substack{\textrm{training}\\
\textrm{set}
}
}\left\Vert N-C\right\Vert ^{2}.\label{eq:E0}
\end{equation}
The normalization factor $n=\text{avg}\left\Vert C\right\Vert $
will be significant when combining different cost functions. To evaluate the results, we average  the normalized root mean squared error across 9 output components
\begin{equation}
e=\frac{1}{9}\sum_{i=1}^{9}\frac{\text{rms}\left(N_{i}-C_{i}\right)}{\sigma_{i}}.\label{eq:value error}
\end{equation}
Here, $\sigma_{i}$ represents the standard deviations of each output component, all approximately
equal to $0.3$.

\subsection{Training Set\label{subsec:training-set}}
The initial cube with a side length of $0.4$ is positioned so that three faces, with intensities $1$, $2/3$, and $1/6$, are equally visible:
\[
C_{0}=\frac{2}{5}\left(\begin{array}{ccc}
\frac{-1}{\sqrt{6}} & 0 & \frac{-1}{\sqrt{6}}\\
\frac{-1}{\sqrt{2}} & 0 & \frac{1}{\sqrt{2}}\\
\frac{-1}{2\sqrt{3}} & \frac{\sqrt{3}}{2} & \frac{-1}{2\sqrt{3}}
\end{array}\dots\right).
\]
Training set is generated by rotating $C_0$ along an arbitrary axis $\vec{r}$ by an angle $\phi \in \left[-\pi/8, \pi/ 8\right]$, followed by a shift with a random vector $\vec{d} \in \left[-0.52, 0.52\right]^3$
\begin{equation}
C\left(\vec{r},\phi,\vec{d}\,\right)=R\left(\vec{r},\phi\right)C_{0}+\vec{d},\label{eq:cube gen}
\end{equation}
where $R$ is the rotation matrix. This avoids the case where only one face is visible and four distinct cube orientations yield the same image. The range of possible variations is represented in Fig.\,\ref{fig:Cube-images}.
The required 6 random parameters are generated using Roberts' version of
the low-discrepancy Korobov sequence \cite{korobov_approximate_1959},
which uniformly covers a 6-dimensional unit cube, as opposed to
a random sequence that occasionally produces regions of higher and lower
density. The first 4 components of the random vector are used for $\vec{d}$ and $\phi$
after a simple rescaling. The last two are scaled to $\rho\in\left[-1,1\right]$,
$\theta\in\left[0,\pi\right]$ and produce $\vec{r}$ as
\[
\begin{cases}
r_{x}=\sqrt{1-\rho^{2}}\cos\theta\\
r_{y}=\sqrt{1-\rho^{2}}\sin\theta\\
r_{z}=\rho
\end{cases}
\]
The training and test sets consist of 97020 and 20160 images respectively.
\subsection{Results\label{subsec:Results}}
To maintain continuity with the previous study \cite{avrutskiy_enhancing_2020}, we use fully connected multilayer perceptrons. Convolutional neural networks, though ubiquitous for many image analysis tasks, are not strictly necessary \cite{tolstikhin_mlp-mixer_2021}.
To estimate the impact of network capacity, we consider neural networks with the following layer configuration
\[
1681,L,L,L,128,9,
\]
where $L=256$, $512$, $1024$, and $2048$. These 4 networks will be referred to by their total number of weights in millions: 0.6, 1.5, 4, and 12, respectively. All hidden layers have activation function
\[
f\left(x\right)=\frac{1}{1+e^{-x}},
\]
the output layer is linear. The weights are initialized with random
values from the range $\pm2/\sqrt{\kappa}$, where $\kappa$ is the
number of senders \cite{glorot_understanding_2010,he_delving_2015}.
For the first matrix, this range is increased to $\pm10/\sqrt{1681}$
to compensate for the input distribution. Thresholds are initialized
in the range $\left[-0.1,0.1\right]$. Training is done with ADAM
\cite{kingma_adam_2014} using the default parameters $\beta_{1}=0.9$,
$\beta_{2}=0.999$, $\epsilon=10^{-8}$. The training set is divided into 42 batches, the number of epochs is 2000. The initial learning rate $l=10^{-3}$ is reduced to $l=10^{-4}$ for the last 50 epochs. Training was performed on Nvidia Tesla A10 using tf32 mode for matrix multiplications.
\begin{table}
\small
\centering
\caption{Error in vertex reconstruction after conventional training\label{tab:accuracy}}
\renewcommand{\arraystretch}{1.3}%
\begin{tabular}{|c|c|c|c|c|c|}
\hline
\multicolumn{2}{|c|}{weights$\times$10\textsuperscript{6}} & 0.6 & 1.5 & 4 & 12\tabularnewline
\hline
\hline
\multirow{2}{*}{$e$,\%} & training & 0.80 & 0.40 & 0.26 & 0.26\tabularnewline
\cline{2-6} \cline{3-6} \cline{4-6} \cline{5-6} \cline{6-6} 
 & test & 1.04 & 0.60 & 0.43 & 0.45\tabularnewline
\hline
\multicolumn{2}{|c|}{time, minutes} & 1.5 & 3 & 6 & 16\tabularnewline
\hline
\end{tabular}
\end{table}
Table $\text{\ref{tab:accuracy}}$ presents the results. The problem does not require significant network capacity: the largest
model shows no improvement over the smaller one. The number of epochs is excessive, as accuracy shows little improvement beyond 500 epochs. However, larger networks and longer training will become essential in later sections; here, these settings are used to maintain consistency.

\subsection{Adversarial Attacks}
\begin{figure}
\centering
\includegraphics[height=3.cm]{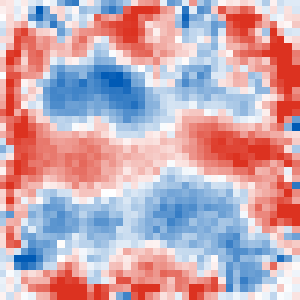}
\includegraphics[height=3.05cm, trim=0.20cm 0.62cm 0.1cm 0.19cm, clip]{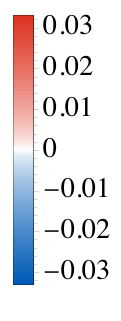}\caption{Gradient of the network with 12 million weights with respect to its first output\label{fig:Adversarial-gradients}}
\end{figure}

The problem ($\text{\ref{eq:problem}}$) and its cost function ($\text{\ref{eq:E0}}$)
are defined only for inputs that lie strictly on the 6-dimensional manifold of the 1681-dimensional image space. The full image space is largely filled with noise, which is of little concern. However, some elements are close enough to the original dataset to appear visually indistinguishable, yet the network's output for them is drastically different \cite{szegedy_intriguing_2013}. As noted
in \cite{goodfellow_explaining_2014}, such images can be generated
using the gradient of the network. An example of
the gradient with respect to the first output is shown in Fig. $\text{\ref{fig:Adversarial-gradients}}$, with a norm of $0.71$. When added to the original image with a small factor
\[
\mathcal{F}=\mathcal{I}\left(C\right)+\varepsilon\nabla N_{1},
\]
the first output can be estimated as
\[
N_{1}\left(\mathcal{F}\right)\simeq N_{1}\left(\mathcal{I}\left(C\right)\right)+\varepsilon\left\Vert \nabla N_{1}\right\Vert ^{2}.
\]
The change is approximately $0.5\varepsilon$, while each individual pixel is altered by only about $0.03\varepsilon$. Even for $\varepsilon\sim1$, this does not change the visual perception
of the cube in the slightest. As a result, two visually identical images
have completely different outputs. This is commonly known as a sensitivity-based
adversarial attack. They generalize well across different inputs, different network capacities, and different training methods \cite{goodfellow_explaining_2014}. This apparent
flaw gave rise to the notion of robustness as the invariance to small perturbations
\cite{qian_survey_2022}. By using the gradient as a tool to estimate the effect of perturbations, researchers made significant progress in training robust networks \cite{lyu_unified_2015,madry_towards_2017}. However,
another attack vector was later identified \cite{jacobsen_excessive_2018}.
Even for networks certified to withstand perturbations up to a certain magnitude, it was possible to modify an image within that range such that the output remained unchanged, but the image itself was fundamentally altered \cite{tramer_fundamental_2020}. This was coined as an invariance-based attack, stemming from the network's lack of response. Robustness was redefined as matching a perfect output (oracle), making invariance-based adversarial detection reliant on human evaluation \cite{jacobsen_excessive_2018,jacobsen_exploiting_2019} and complicating both analysis and training. For the problem at hand, the data manifold is computationally accessible, allowing us to construct an oracle and explore all aspects of robustness numerically.

\subsection{Oracle}
The oracle is a function that matches ($\text{\ref{eq:problem}}$)
on manifold images and returns ``the best possible output'' for any
other image $\mathcal{F}$. We define it as the cube whose image is closest in Euclidean distance
\begin{equation}
\mathcal{O}\left(\mathcal{F}\right)=\arg\min_{C}\left\Vert \mathcal{I}\left(C\right)-\mathcal{F}\right\Vert ^{2}.\label{eq:extended problem}
\end{equation}
For an arbitrary $\mathcal{F}$ from 1681-dimensional image space, this leads to a nonlinear 6-dimensional minimization problem. Solving it directly would require rendering an image for each iteration. However, if $\mathcal{F}$ is close to the manifold
$\mathcal F=\mathcal I(C) +\varepsilon\mathcal D$, the oracle can be evaluated using a Taylor expansion
\begin{equation}
\mathcal{O}\left(\mathcal{I}+\varepsilon\mathcal{D}\right)=C+\varepsilon\frac{\partial\mathcal{O}}{\partial\mathcal{D}}+\frac{\varepsilon^{2}}{2}\frac{\partial^{2}\mathcal{O}}{\partial\mathcal{D}^{2}}+o\left(\varepsilon^{2}\right).\label{eq:oracle expansion}
\end{equation}
Its terms can be computed from image derivatives with respect cube's degrees of freedom. The robustness of the network can be analyzed using the Taylor expansion of
\begin{equation}
\left\Vert N\left(\mathcal{F}\right)-\mathcal{O}\left(\mathcal{F}\right)\right\Vert ^{2}.\label{eq:N-O}
\end{equation}

\section{First Order\label{sec:first-order}}

This section focuses on first-order analysis, beginning with the calculation of image derivatives. Oracle expansion is discussed in Subsection \ref{subsec:oracle-d1}, and the robustness analysis is covered in Subsection \ref{subsec:Analysis-of-robustness}.
\subsection{Local Tangent Space \label{subsec:tangent space}}
\begin{figure}
\centering
\subfloat[\label{fig:subX}]{
\includegraphics[height=3.0cm]{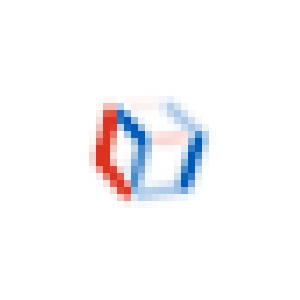}}
\includegraphics[height=2.8cm, trim=0.09cm 0.6cm 0.09cm 0.15cm, clip]{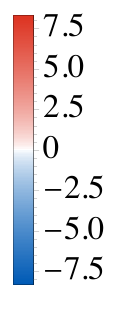}
\subfloat[\label{fig:subZ}]{\includegraphics[height=3.0cm]{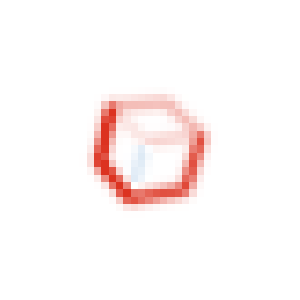}

}\includegraphics[height=2.8cm, trim=0.09cm 0.6cm 0.09cm 0.15cm, clip]{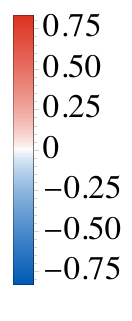}

\subfloat[\label{fig:rX}]{\includegraphics[height=3.0cm]{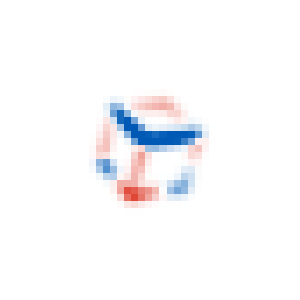}

}\includegraphics[height=2.8cm, trim=0.09cm 0.6cm 0.09cm 0.15cm, clip]{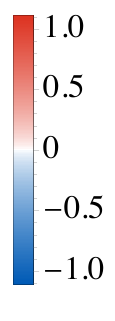}\subfloat[\label{fig:rZ}]{\includegraphics[height=3.0cm]{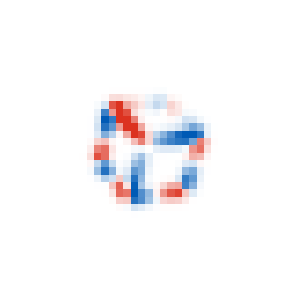}

}\includegraphics[height=2.8cm, trim=0.09cm 0.6cm 0.09cm 0.15cm, clip]{img2/1.0a.pdf}

\caption{First-order image derivatives corresponding to infinitesimal operators:
translations along the $x$ (a) and $z$ (b) axes, and rotations
about the $x$ (c) and $z$ (d) axes\label{fig:Image-d1}}
\end{figure}

In linear approximation, the data manifold near each cube is represented by a hyperplane spanned by 6 independent tangent vectors. Those vectors can be obtained from the image derivatives with respect to cube's degrees of freedom. Cube's parametrization (\ref{eq:cube gen}) by $\vec{r}$, $\phi$ and $\vec{d}$ yields
\[
\frac{\partial\mathcal{I}}{\partial r_{i}},\frac{\partial\mathcal{I}}{\partial\phi},\frac{\partial\mathcal{I}}{\partial d_{j}},
\]
where $i=1,2$ and $j=1,2,3$. However, if $\phi=0$, both derivatives with respect to $r_{i}$ are zero, and a 6D basis cannot be formed. This issue, common to any set of global and continuous parameters determining the orientation of a rigid body, arises from the intrinsic properties of the rotational group \cite{hemingway_perspectives_2018}.
To address this, a local parametrization will be used. Consider a cube $C$ and three rotations $R_X$, $R_Y$ and $R_Z$ at angles $\nu_1$, $\nu_2$ and $\nu_3$ about the basis $X$, $Y$ and $Z$-axes, respectively, all passing through the center of the cube. Their composition\footnote{note that $R_{X}R_{Y}\ne R_{Y}R_{X}$, etc}
$R_{Z\!}\left(\nu_{3}\right)R_{Y\!}\left(\nu_{2}\right)R_{X\!}\left(\nu_{1}\right)$
has a fixed point at $\nu_{1}=\nu_{2}=\nu_{3}=0$, and is continuously differentiable in the neighbourhood of this point. Combined with three translations along the same axes by $\nu_4$, $\nu_5$ and $\nu_6$ respectively, this yields a locally smooth transformation with 6 independent parameters. For a vertex $(x,y,z)$, it is expressed as
\begin{equation}
\begin{aligned}
\left(\begin{array}{c}
\!\!x'\!\!\\
\!\!y'\!\!\\
\!\!z'\!\!
\end{array}\right)\!= \! \left(\begin{array}{c}
\!\!d_{1}+\nu_{4}\!\!\\
\!\!d_{2}+\nu_{5}\!\!\\
\!\!d_{3}+\nu_{6}\!\!
\end{array}\right)\!+
R_{Z}\!\left(\nu_{3}\right)R_{Y}\!\left(\nu_{2}\right)R_{X}\!\left(\nu_{1}\right)\!\! &\left(\begin{array}{c}
\!\!x-d_{1}\!\!\\
\!\!y-d_{2}\!\!\\
\!\!z-d_{3}\!\!
\end{array}\right) \! \!,\label{eq:T}
\end{aligned}
\end{equation}
where $\vec{d}$ is the center of mass of the cube. This transformation applied to each vertex of the cube is denoted as
\begin{equation}
C'=\hat{T}\left(\nu_j\right)C,\quad  j\in[1,6]	.
\end{equation}
The differential operators $\partial \hat{T}/\partial \nu_j$ are linearly independent at $\nu_j=0$. They
represent 6 infinitesimal motions: three rotations and three translations. The image derivatives are defined as
\begin{equation}
\frac{\partial\mathcal{I}}{\partial\nu_{i}}=\lim_{\varepsilon\rightarrow0}\frac{\mathcal{I}\left(\hat{T}\left(\varepsilon\delta_{ij}\right)C\right)-\mathcal{I}\left(C\right)}{\varepsilon},\label{eq:dI/dv}
\end{equation}
where $\varepsilon\delta_{ji}$ denotes a vector with index $j$ where $\varepsilon$ is its $i$-th component and all other components are zero. Examples are shown in Fig. $\text{\ref{fig:Image-d1}}$. By adding
these images to the original one, the cube can be moved within $\pm 10\%$
of its length in the $x$ and $y$ directions, within $\pm 75\%$ in $z$ direction,
and rotated by up to 10 degrees about any axis. Details on the calculation of image derivatives are provided in Appendix~\ref{Derivatives-of-images}.

Thus, in the vicinity of any point on the manifold, we obtain first-order expansions for the cube's vertices and image
\begin{equation}
\hat{T}\left(\nu_{i}\right)C=C+\frac{\partial C}{\partial\nu_{i}}\nu_{i}+o\left(\nu_{i}\right),\label{eq:cube first order}
\end{equation}
\begin{equation}
\mathcal{I}\left(\hat{T}\left(\nu_{i}\right)C\right)=\mathcal{I}\left(C\right)+\frac{\partial\mathcal{I}}{\partial\nu_{i}}\nu_{i}+o\left(\nu_{i}\right).\label{eq:first order series}
\end{equation}

\subsection{Improving Accuracy\label{subsec:accuracy-1}}

Training derivatives can significantly improve the network's accuracy, as established in \cite{avrutskiy_enhancing_2020}. The forward pass procedure is extended and supplied with input derivatives, which allow for the calculation of the corresponding output derivatives
\[
\mathcal{I},\frac{\partial\mathcal{I}}{\partial\nu_{i}}\overset{\substack{\textrm{forward}\\
\textrm{pass}
}
}{\longrightarrow}N,\frac{\partial N}{\partial\nu_{i}},
\]
used in the additional first-order cost
\[
E_{1}=\sum_{\substack{\textrm{training}\\
\textrm{set}
}
}\left[\sum_{i=1}^{6}\frac{1}{n_{i}^{2}}\left\Vert \frac{\partial N}{\partial\nu_{i}}-\frac{\partial C}{\partial\nu_{i}}\right\Vert ^{2}\right],
\]
added to the conventional cost function (\ref{eq:E0}). The normalization factor $n_i$ is the average norm of the output derivative, $n_{i}=\text{avg}\left\Vert \partial C/\partial\nu_{i}\right\Vert $.
The extended backward pass calculates the gradient of $E=E_{0}+E_{1}$ with respect to the weights. Training parameters are the same as those in Subsection \ref{subsec:Results}.
\begin{table}
\small
\caption{
Error in vertex reconstruction after first order training\label{tab:accuracy-d1}}
\centering
\renewcommand{\arraystretch}{1.3}%
\begin{tabular}{|c|c|c|c|c|c|}
\hline
\multicolumn{2}{|c|}{weights$\times$10\textsuperscript{6}} & 0.6 & 1.5 & 4 & 12 \tabularnewline
\hline
\hline
\multirow{2}{*}{$e$,\%} & training & 0.16 & 0.11 & 0.067 & 0.065\tabularnewline
\cline{2-6} \cline{3-6} \cline{4-6} \cline{5-6} \cline{6-6} 
 & test & 0.17 & 0.12 & 0.075 & 0.072\tabularnewline
\hline 
\end{tabular}
\end{table}
The results are shown in Table $\text{\ref{tab:accuracy-d1}}$.
Compared to the conventional training (Table \ref{tab:accuracy}), the accuracy increases by a factor of 5, similar to the improvements observed in low-dimensional applications. Overfitting is largely resolved, even for larger models, confirming the results from \cite{avrutskiy_preventing_2020}. Due to the excessive number of epochs, it is highly unlikely that the accuracy gap between conventional and proposed training can be bridged. The increase in training time can be estimated by multiplying the times in Table \ref{tab:accuracy} by 6; however, this can be significantly reduced \cite{avrutskiy_enhancing_2020}. The network's gradients with respect to the outputs are similar to those from conventional training (see Fig. \ref{fig:Adversarial-gradients}), so a perturbation with norm $0.02$ can negate most improvements. Overall, this training is better suited for image generation, which involves a low-dimensional input and a high-dimensional output, or for autoencoders.

\subsection{Oracle Expansion\label{subsec:oracle-d1}}
The accuracy across the entire input space, and therefore the resilience to perturbations, is defined by the deviation from the oracle (\ref{eq:N-O}). Its linear analysis near the manifold requires computing the gradients of $N$ and $\mathcal O$. While the gradient of $N$ can be efficiently computed using the backward pass, the gradient of the oracle requires first determining the directional derivative from the expression (\ref{eq:oracle expansion}).
\subsubsection{Derivative}
To find the first-order directional derivative
\begin{equation}
\frac{\partial\mathcal{O}}{\partial\mathcal{D}}=\lim_{\varepsilon\rightarrow0}\frac{\mathcal{O}\left(\mathcal{I}\left(C\right)+\varepsilon\mathcal{D}\right)-\mathcal{O}\left(\mathcal{I}\left(C\right)\right)}{\varepsilon},\label{eq:ddir}
\end{equation}
we must determine the term $\mathcal{O}\left(\mathcal{I}\left(C\right)+\varepsilon\mathcal{D}\right)$.
Since any output of the oracle is a cube, the result can be written
as a transformation of the initial one
\begin{equation}
\mathcal{O}\left(\mathcal{I}\left(C\right)+\varepsilon\mathcal{D}\right)=\hat{T}\left(\nu_{i}\left(\varepsilon\right)\right)C,\label{eq:oracle 1d}
\end{equation}
where $\nu_{i}\left(\varepsilon\right)$ are some unknown functions, $\nu_{i}\left(0\right)=0$.
Substituting this and $\mathcal{F}=\mathcal{I}\left(C\right)+\varepsilon\mathcal{D}$
into the oracle definition ($\text{\ref{eq:extended problem}}$),
we obtain a minimization problem for $\nu_{i}\left(\varepsilon\right)$
\begin{equation}\label{eq:interm}
\min_{\nu_{i}\left(\varepsilon\right)}\left\Vert \mathcal{I}\left(\hat{T}\left(\nu_{i}\left(\varepsilon\right)\right)C\right)-\mathcal{I}\left(\mathcal{C}\right)-\varepsilon\mathcal{D}\right\Vert ^{2}.
\end{equation}
Its solution can be obtained using a Taylor expansion
\[
\nu_{i}\left(\varepsilon\right)=\varepsilon\frac{d\nu_{i}}{d\varepsilon}+o\left(\varepsilon\right).
\]
Plugging it and the first-order expansion of images ($\text{\ref{eq:first order series}}$) into (\ref{eq:interm})
we obtain a minimization problem for $d\nu_{i}/d\varepsilon$
\[
\min_{d\nu_{i}/d\varepsilon}\left\Vert \frac{\partial\mathcal{I}}{\partial\nu_{i}}\frac{d\nu_{i}}{d\varepsilon}-\mathcal{D}\right\Vert ^{2}.
\]
By expanding the squared norm and differentiating with respect
to $d\nu_{j}/d\varepsilon$, we get
\[
\biggl(\underbrace{\frac{\partial\mathcal{I}}{\partial\nu_{i}}\cdot\frac{\partial\mathcal{I}}{\partial\nu_{j}}}_{\Gamma_{ij}}\biggr)\frac{d\nu_{i}}{d\varepsilon}-\frac{\partial\mathcal{I}}{\partial\nu_{j}}\cdot\mathcal{D}=0.
\]
The expression in parentheses denotes the $6 \times 6$ Gram matrix of the tangent vectors. The resulting equation is:
\begin{equation}
\Gamma_{ij}\frac{d\nu_{i}}{d\varepsilon}=\mathcal{D}\cdot\frac{\partial\mathcal{I}}{\partial\nu_{j}},\label{eq:first directional}
\end{equation}
and it can be solved by computing $\Gamma^{-1}$, which exists if the
vectors $\partial\mathcal{I}/\partial\nu_{j}$ are linearly independent,
a condition satisfied at any non-special point of the 6-dimensional hypersurface. The output of the oracle is
\[
\mathcal{O}\left(\mathcal{I}\left(C\right)+\varepsilon\mathcal{D}\right)=C+\varepsilon\frac{\partial C}{\partial\nu_{i}}\frac{d\nu_{i}}{d\varepsilon},
\]
and thus the first derivative is
\[
\frac{\partial\mathcal{O}}{\partial\mathcal{D}}=\frac{\partial C}{\partial\nu_{i}}\frac{d\nu_{i}}{d\varepsilon},
\]
where $d\nu_i/d \varepsilon$ is the solution of (\ref{eq:first directional}).

\subsubsection{Gradient\label{subsec:Oracle grad}}

We first determine the gradient direction: a normalized vector $\mathcal D$ along which a given output increases most rapidly. For example, we choose the first output $\mathcal{O}_{1}$, which is $C_{1}$: the $x$-coordinate of the first vertex of the cube. From ($\text{\ref{eq:first directional}}$) it follows that only the projection of $\mathcal{D}$ onto the tangent vectors $\partial\mathcal{I}/\partial\nu_{j}$
can contribute to $d\nu_{i}/d\varepsilon$ and thus to any change
in $\mathcal{O}_{1}$. For $\mathcal D$ to be an extremum, it must
lie entirely in the 6-dimensional tangent space, and can thus be represented
by a vector $\nu_{i}$:
\begin{equation}
\mathcal{D}=\frac{\partial\mathcal{I}}{\partial\nu_{i}}\nu_{i}.\label{eq:oracle grad}
\end{equation}
This implies that the gradient of the oracle always corresponds to a valid cube motion, in which case the equation ($\text{\ref{eq:first directional}}$) has a simple solution $d\nu_{i}/d\varepsilon=\nu_{i}$, and therefore
\[
\mathcal{O}\left(\mathcal{I}\left(C\right)+\varepsilon\mathcal{D}\right)=C+\varepsilon\frac{\partial C}{\partial\nu_{i}}\nu_{i}.
\]
To find $\nu_i$ we need to maximize the first component subject to the constraint on the norm of $\mathcal D$:
\[
\begin{cases}
\max_{\nu_{i}}\dfrac{\partial C_{1}}{\partial\nu_{i}}\nu_{i}\\
\left\Vert \mathcal{D}\right\Vert ^{2}=\Gamma_{ij}\nu_{i}\nu_{j}=1.
\end{cases}
\]
By writing the Lagrangian and taking its partial derivatives, we obtain
\[
\begin{cases}
2\lambda\Gamma_{ij}\nu_{j}+\dfrac{\partial C_{1}}{\partial\nu_{i}}=0\\
\Gamma_{ij}\nu_{i}\nu_{j}=1.
\end{cases}
\]
The first equation can be solved for a single $\lambda$, yielding $\nu_i$,
which is then scaled to satisfy the second equation. Assuming $\lambda=-1/2$ 
\begin{equation}
\Gamma_{ij}\nu_{j}=\dfrac{\partial C_{1}}{\partial\nu_{i}}.\label{eq:or grad eq}
\end{equation}
This equation can be used to reduce the norm to:
\[
\left\Vert \mathcal{D}\right\Vert ^{2} = \frac{\partial C_{1}}{\partial\nu_{i}}\nu_{i},
\]
allowing the normalized gradient to be written as
\begin{equation}
\mathcal{D}=\frac{\frac{\partial\mathcal{I}}{\partial\nu_{i}}\nu_{i}}{\sqrt{\frac{\partial\mathcal{C}_{1}}{\partial\nu_{i}}\nu_{i}}}\equiv\frac{1}{\sqrt{g}}\frac{\partial\mathcal{I}}{\partial\nu_{i}}\nu_{i}.\label{eq:d-norm}
\end{equation}
The magnitude of the gradient can be determined from the directional derivative:
\begin{align*}
\left\Vert \nabla\mathcal{O}_{1}\right\Vert  & =\frac{\partial\mathcal{O}_{1}}{\partial\mathcal{D}}=\\
 & =\lim_{\varepsilon\rightarrow0}\frac{1}{\varepsilon}\left(\mathcal{O}_{1}\left(\mathcal{I}+\frac{\varepsilon}{\sqrt{g}}\frac{\partial\mathcal{I}}{\partial\nu_{i}}\nu_{i}\right)-\mathcal{O}_{1}\left(\mathcal{I}\right)\right)=\\
 & =\frac{1}{\sqrt{g}}\frac{\partial C_{1}}{\partial\nu_{i}}\nu_{i}=\sqrt{g}.
\end{align*}
This cancels the normalization factor in ($\ref{eq:d-norm}$), so the resulting gradient of the oracle is simply
\[
\nabla\mathcal{O}_{1}=\frac{\partial\mathcal{I}}{\partial\nu_{j}}\nu_{j},
\]
where $\nu_{i}$ is the solution of ($\text{\ref{eq:or grad eq}}$).
An example is shown in Fig. \ref{fig:g_oracle}. A total of 9 gradients
belong to the 6-dimensional tangent space, so they are not linearly independent.

\subsection{Robustness Analysis\label{subsec:Analysis-of-robustness}}
Before discussing optimal adversarial attacks, we first note a simple observation. Consider the first component of the error $N_{1}-\mathcal{O}_{1}$. Whenever the network and the oracle have unequal gradients, $\nabla N_{1}\neq\nabla\mathcal{O}_{1}$, we can subtract the projection of $\nabla N_1$ onto $\nabla \mathcal{O}_1$ from $\nabla \mathcal{O}_1$ or vice versa, yielding two possible directions
\begin{equation}
\mathcal{D}_{\text{inv.}}=\nabla\mathcal{O}_{1}-\left(\nabla N_{1}\cdot\nabla\mathcal{O}_{1}\right)\frac{\nabla N_{1}}{\left\Vert \nabla N_{1}\right\Vert ^{2}},\label{eq:inv_simp}
\end{equation}
\begin{equation}
\mathcal{D}_{\text{sen.}}=\nabla N_{1}-\left(\nabla N_{1}\cdot\nabla\mathcal{O}_{1}\right)\frac{\nabla\mathcal{O}_{1}}{\left\Vert \nabla\mathcal{O}_{1}\right\Vert ^{2}}.\label{eq:sen_simp}
\end{equation}
Moving along $\mathcal{D}_{\text{inv.}}$, the network output $N_{1}$
remains constant, while the oracle output $\mathcal{O}_{1}$ changes, representing an invariance-based adversarial attack. Conversely, $\mathcal{D}_{\text{sen.}}$ corresponds to a sensitivity-based attack. Both vanish if the gradients are equal, and there is no trade-off of any kind. Without robust training, the term $\left(\nabla N_{1}\cdot\nabla\mathcal{O}_{1}\right)$ is typically negligible, making $\mathcal{D}_{\text{sen.}}$ essentially an adversarial gradient. Minimizing the network's response in this direction improves robustness; however, during training, the network's gradient eventually becomes ``perceptually aligned''
\cite{santurkar_image_2019,kaur_are_2019}, at which point the term $\left(\nabla N_{1}\cdot\nabla\mathcal{O}_{1}\right)$ is no longer small, so further training does not improve robustness.

\subsubsection{Optimal Attack}
We seek to construct the optimal adversarial attack: a bounded
perturbation that maximizes the error,
\begin{equation}
\begin{cases}
\max_{\mathcal{D}}\left\Vert N_i\left(\mathcal{I}+\varepsilon\mathcal{D}\right)-\mathcal{O}_i\left(\mathcal{I}+\varepsilon\mathcal{D}\right)\right\Vert ^{2}\\
\left\Vert \mathcal{D}\right\Vert ^{2}\leq1.
\end{cases}\label{eq:univ. p}
\end{equation}
Here, $\varepsilon$ should be small enough for the first-order approximations
to hold, in which case
\begin{equation}
\begin{split}
&N_{i}\left(\mathcal{I}+\varepsilon\mathcal{D}\right)-\mathcal{O}_{i}\left(\mathcal{I}+\varepsilon\mathcal{D}\right)=\\=&\;N_{i}\left(\mathcal{I}\right)-\mathcal{O}_{i}\left(\mathcal{I}\right)+\varepsilon\mathcal{D}\cdot\nabla\left(N_{i}-\mathcal{O}_{i}\right).\label{eq:1order dif}
\end{split}
\end{equation}
We are most interested in the scenario where the perturbation significantly increases the initial error
\begin{equation}
\varepsilon\left\Vert \mathcal{D}\cdot\nabla\left(N_{i}-\mathcal{O}_{i}\right)\right\Vert \gg\left\Vert N_{i}\left(\mathcal{I}\right)-\mathcal{O}_{i}\left(\mathcal{I}\right)\right\Vert .\label{eq:assumption}
\end{equation}
With this condition, the problem is decoupled from $\varepsilon$ and the
constraint can be replaced by the equality
\begin{equation}
\begin{cases}
\max_{\mathcal{D}}\left\Vert \mathcal{D}\cdot\nabla\left(N_{i}-\mathcal{O}_{i}\right)\right\Vert ^{2}\\
\left\Vert \mathcal{D}\right\Vert ^{2}=1.
\end{cases}\label{eq:optimal}
\end{equation}
Thus, $\mathcal{D}$ is the direction in which the deviation from the oracle
increases most rapidly. As long as this maximum is large enough for
(\ref{eq:assumption}) to hold with an acceptable $\varepsilon$,
we do not need to consider the full expression (\ref{eq:1order dif}).
This simplification has proven to be appropriate for all cases considered.
The magnitude of the maximum in (\ref{eq:optimal}) can be used as a
measure of the network's vulnerability after appropriate normalization.
Since the standard deviations of all outputs are approximately 0.3, a reasonable
metric is
\begin{equation}
v=\frac{1}{0.3}\frac{\sqrt{\max}}{9},\label{eq:vuner}
\end{equation}
which represents the rate at which the relative error increases as we move along the optimal attack vector. We will denote it by $v_{\text{max}}$ as the maximum vulnerability. The optimal attack vector is a linear combination of the gradients of the difference
\[
\mathcal{D}=\mu_{i}\nabla\left(N_{i}-\mathcal{O}_{i}\right),
\]
by substituting it into (\ref{eq:optimal}) and denoting
\[
G_{ij}=\nabla\left(N_{i}-\mathcal{O}_{i}\right)\cdot\nabla\left(N_{j}-\mathcal{O}_{j}\right),
\]
we obtain a constrained maximization problem
\begin{equation}
\begin{cases}
\max_{\mu}\left\Vert \mu_{i}G_{ij}\right\Vert ^{2}\\
\mu_{i}\mu_{j}G_{ij}=1,
\end{cases}\label{eq:max vun}
\end{equation}
the solution to which is the eigenvector of $G_{ij}$ corresponding to the maximum eigenvalue
\[
\mu_{i}G_{ij}=\lambda_{\max}\mu_{j},
\]
and this eigenvalue is the magnitude of the maximum
\[
\max_{\mu}\left\Vert \mu_{i}G_{ij}\right\Vert ^{2}=\lambda_{\max}.
\]
\subsubsection{Invariance and Sensitivity-based Attacks}
The construction of optimal attacks that exploit solely invariance or sensitivity is more nuanced. The optimal vector can be expressed as a linear combination of gradients
\begin{equation}
\mathcal{D}=\nu_{i}\nabla\mathcal{O}_{i}+\omega_{i}\nabla N_{i},\label{eq:Jnb}
\end{equation}
where $i\in\left[1,9\right]$, making this an 18-dimensional problem.
However, only 6 components of $\nabla\mathcal{O}_{i}$ are linearly independent, and the situation for $\nabla N_{i}$ is very similar: the behavior of $N$ is well described in terms of the 6 largest eigenvectors
of the matrix $\nabla N_{i}\cdot\nabla N_{j}$. For clean images,
this arises from having only 6 degrees of freedom, although it also holds remarkably well for noisy and corrupted images. Denoting
{\renewcommand{\arraystretch}{1.3}
\begin{equation}
\left(\begin{array}{cc}
G_{ij}^{11} & G_{ij}^{12}\\
G_{ij}^{21} & G_{ij}^{22}
\end{array}\right)=\left(\begin{array}{cc}
\nabla\mathcal{O}_{i}\cdot\nabla\mathcal{O}_{j} & \nabla\mathcal{O}_{i}\cdot\nabla N_{j}\\
\nabla N_{i}\cdot\nabla\mathcal{O}_{j} & \nabla N_{i}\cdot\nabla N_{j}
\end{array}\right),\label{eq:Gij}
\end{equation}
}
and substituting ($\text{\ref{eq:Jnb}}$) into (\ref{eq:optimal})
we get
\begin{equation}
\begin{cases}
\max_{\nu,\omega}\left\Vert \nu_{i}\left(G_{ij}^{12}-G_{ij}^{11}\right)+\omega_{i}\left(G_{ij}^{22}-G_{ij}^{21}\right)\right\Vert ^{2}\\
\nu_{i}\nu_{j}G_{ij}^{11}+\omega_{i}\omega_{j}G_{ij}^{22}+2\nu_{i}\omega_{j}G_{ij}^{12}=1.
\end{cases}\label{eq:spec attack}
\end{equation}
An additional constraint is required to specify the type of attack.
For a sensitivity-based attack, it is $\mathcal{D}_{\text{sen.}}\cdot \nabla \mathcal{O}_{i}=0$, or
\begin{equation}
\nu_{i}G_{ij}^{11}+\omega_{i}G_{ij}^{21}=0.\label{eq:pert}
\end{equation}
This relation eliminates 2 out of the 4 terms in the expression to
be maximized. It consists of 9 equations, whereas we expect 6. The extra equations
can be removed by projecting (\ref{eq:pert}) onto the 6 eigenvectors of
$G^{22}$ with largest eigenvalues. The optimization problem
for sensitivity-based attack is
\begin{equation}
\begin{cases}
\max_{\nu,\omega}\left\Vert \nu_{i}G_{ij}^{12}+\omega_{i}G_{ij}^{22}\right\Vert ^{2}\\
\nu_{i}\nu_{j}G_{ij}^{11}+\omega_{i}\omega_{j}G_{ij}^{22}+2\nu_{i}\omega_{j}G_{ij}^{12}=1\\
a_{kj}\left(\nu_{i}G_{ij}^{11}+\omega_{i}G_{ij}^{21}\right)=0\\
a_{ki}G_{ij}^{22}=\lambda_{k}a_{kj},\;k\in\left[1,6\right]
\end{cases}\label{eq:sen}
\end{equation}
In terms of the stacked vector $\left(\nu_{i},\omega_{i}\right)$ the first
constraint is a quadratic form with the matrix (\ref{eq:Gij}). The expression
to be maximized is also a quadratic form. Denoting $P=G_{ij}^{12}$ and
$W=G_{ij}^{22}$, its matrix is
\begin{equation}
\left(\begin{array}{cc}
PP^{T} & PW^{T}\\
WP^{T} & WW^{T}
\end{array}\right).\label{eq:PW}
\end{equation}
The solution method is as follows: First, we find a basis in which
the matrix (\ref{eq:Gij}) becomes the identity. In this basis, we
express the general solution to the second constraint as a linear combination
of orthonormal vectors. These vectors form a new basis, in
which we compute the largest eigenvalue of the matrix (\ref{eq:PW}), which gives the desired maximum. For invariance attack, the procedure is similar; however, instead of equation (\ref{eq:pert}) we have
\[
\nu_{i}G_{ij}^{12}+\omega_{i}G_{ij}^{22}=0,
\]
and the optimization problem is
\begin{equation}
\begin{cases}
\max_{\nu,\omega}\left\Vert \nu_{i}G_{ij}^{11}+\omega_{i}G_{ij}^{21}\right\Vert ^{2}\\
\nu_{i}\nu_{j}G_{ij}^{11}+\omega_{i}\omega_{j}G_{ij}^{22}+2\nu_{i}\omega_{j}G_{ij}^{12}=1\\
a_{kj}\left(\nu_{i}G_{ij}^{12}+\omega_{i}G_{ij}^{22}\right)=0\\
a_{ki}G_{ij}^{11}=\lambda_{k}a_{kj},\;k\in\left[1,6\right]
\end{cases}\label{eq:inv}
\end{equation}
The solution is the same, except that the elements of (\ref{eq:PW})
are $P=G_{ij}^{11}$ and $W=G_{ij}^{21}$. The network vulnerability is quantified by the obtained maxima using the formula (\ref{eq:vuner}), with the results denoted by $v_{\text{sen.}}$ for sensitivity-based attack and $v_{\text{inv.}}$ for invariance-based attack.

\subsection{Universal Robust Training\label{subsec:Robust-training}}

As prescribed by linear theory, the endpoint of robust training is a network
whose gradient $\nabla N$ matches the gradient of the oracle $\nabla\mathcal{O}$.
Although there is an efficient method to compute $\nabla N$ using
the backward pass, training it lacks a similarly simple approach. One could construct
a new computational graph that includes both forward and backward passes,
but in this case, each weight would have a double contribution, which
creates multiple paths and significantly increases the computational complexity for
backpropagation. This challenge can also be approached from a different perspective. The condition imposed on the gradient is equivalent to conditions on the output's derivatives with respect to all inputs, as these derivatives can be computed from the gradient. Therefore, we expect the gradient alignment problem to be comparable to training all 1681 derivatives across all input patterns, motivating our aim to minimize
\begin{equation}
\left\Vert \frac{\partial N}{\partial\mathcal{D}}-\frac{\partial\mathcal{O}}{\partial\mathcal{D}}\right\Vert ^{2}\label{eq:rob_g}
\end{equation}
for any image on the manifold and any direction $\left\Vert \mathcal{D}\right\Vert =1$. This can be achieved by generating random directions \cite{avrutskiy_neural_2020} and shuffling them after each epoch. This offloads the number of directions per pattern onto the number of epochs and is the primary reason for the increase in iterations compared to the non-robust formulation.

\subsubsection{Random Directions}
\begin{figure}
\centering
\subfloat[\label{fig:J1}]{\includegraphics[height=2.6cm]{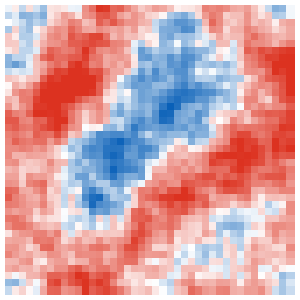}}
\subfloat[\label{fig:J2}]{\includegraphics[height=2.6cm]{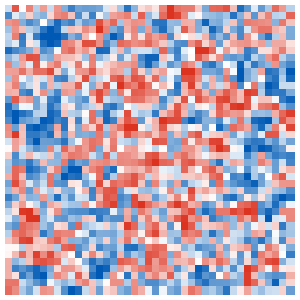}}
\subfloat[\label{fig:J3}]{\includegraphics[height=2.6cm]{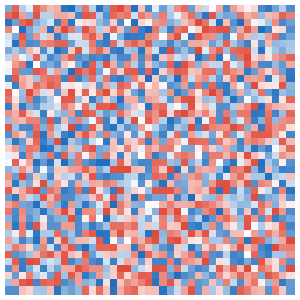}}
\includegraphics[height=2.6cm, trim=0.17cm 0.50cm 0.19cm 0.2cm, clip]{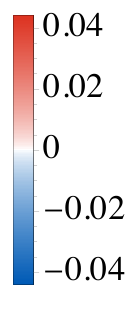}

\caption{Three classes of random directions of the image space}
\end{figure}

To obtain a representative set of random directions, we use three classes of images with different spatial properties. The first mimics the properties of adversarial gradients of the original problem. The second and third are created to address the remaining misalignments with oracle's gradients. As a building block, we use the convolution of white noise with a Gaussian matrix
\[
G_{ij}\left(r\right)=\frac{1}{n}\exp\left(\frac{-2}{r^{2}}\left[\left(i-r-1\right)^{2}+\left(j-r-1\right)^{2}\right]\right),
\]
where $r$ is the blur radius, $i,j\in\left[1,2r+1\right]$, and $n$ is the normalizing factor. By convolving it with the white noise matrix $U$ of size $41+2r$ and normalizing the result, we get
\[
\mathcal{U}\left(r\right)=U\circledast G\left(r\right) / \left\Vert \,U\circledast G\left(r\right)\right\Vert ,
\]
an image in which frequencies above approximately $1/r$ are filtered. Images similar to the adversarial gradient shown in Fig. \ref{fig:Adversarial-gradients} can be generated as
\[
\mathcal{D}_{\text{low}}=\mathcal{U}\!\left(15\right)+\mathcal{U}\!\left(8\right)+0.4\,\mathcal{U}\!\left(2\right)+0.2\,\mathcal{U}\!\left(1\right),
\]
followed by normalization $\mathcal{D}_{\text{low}}=\mathcal{D}_{\text{low}}/\left\Vert \mathcal{D}_{\text{low}}\right\Vert$, see Fig. \ref{fig:J1}. Note that adversarial gradients cannot be used directly, as they are too similar across the training samples.

Minimizing (\ref{eq:rob_g}) using $\mathcal{D}_{\text{low}}$ completely
transforms the network gradients; however, a significant misalignment still persists, dominated by higher frequencies. To address this, we generate a second set of directions
\[
\mathcal{D}_{\text{mid}}=\left[\, \mathcal{U}\!\left(2\right)+\mathcal{U}\!\left(1\right) \right]/ \left\Vert \,\mathcal{U}\!\left(2\right)+\mathcal{U}\!\left(1\right)\right\Vert ,
\]
an example is shown in Fig. \ref{fig:J2}. Finally the third set of directions is unfiltered white noise
\[
\mathcal{D}_{\text{high}}=\mathcal{U}\!\left(0\right),
\]
see Fig. \ref{fig:J3}. Including mid- and high-frequency directions reduces the cost function for low frequencies, but the reverse is not true. Using only high- and mid-frequency components results in noticeable low-frequency misalignment with the oracle's gradients.

\subsubsection{Results}

\begin{figure}
\centering
\subfloat[\label{fig:g_oracle}]{\includegraphics[height=2.5cm, trim=0.08cm 0.08cm 0.08cm 0.08cm, clip]{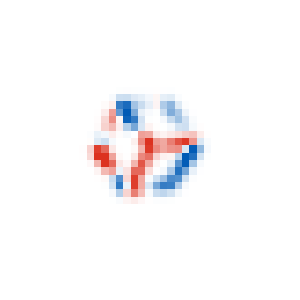}
}\subfloat[\label{fig:g_robust}]{\includegraphics[height=2.5cm, trim=0.08cm 0.08cm 0.08cm 0.08cm, clip]{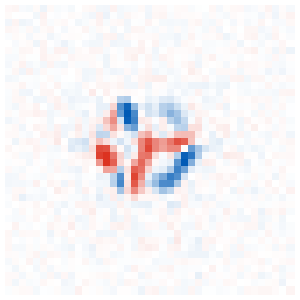}
}\subfloat[\label{fig:g_gauss}]{\includegraphics[height=2.5cm, trim=0.08cm 0.08cm 0.08cm 0.08cm, clip]{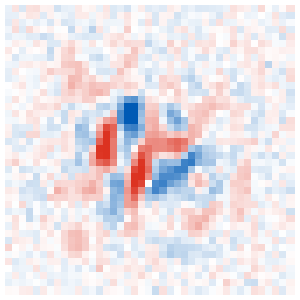}

}\includegraphics[height=2.55cm, trim=0.17cm 0.5cm 0.19cm 0.19cm, clip]{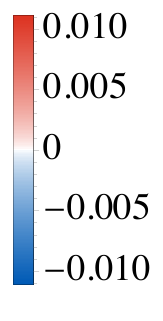}

\caption{Gradients with respect to the first output for: (a) the oracle, (b) the network trained with the first-order universal robust algorithm, and (c) the network trained with Gaussian augmentation.\label{fig:Gradient-oracle}}
\end{figure}

The expression for first-order universal robust training using $\mathcal{D}_{\text{low}}$ is given by
\begin{equation}
R_{\text{low}}^{I}=\frac{1}{c^{2}}\sum_{\substack{\textrm{training}\\
\textrm{set}
}
}\left[\sum_{i=1}^{9}\frac{1}{n_{i}^{2}}\left(\frac{\partial N_{i}}{\partial\mathcal{D}_{\text{low}}}-\frac{\partial\mathcal{O}_{i}}{\partial\mathcal{D}_{\text{low}}}\right)^{2}\right].\label{eq:R1low}
\end{equation}
The normalization of each output component is based on the range between the 95th and 5th percentiles of the target
\[
n_{i}=P_{95}\left[\frac{\partial\mathcal{O}_{i}}{\partial\mathcal{D}_{\text{low}}}\right]-P_{5}\left[\frac{\partial\mathcal{O}_{i}}{\partial\mathcal{D}_{\text{low}}}\right],
\]
and the factor $c$ normalizes the average norm of the result
\[
c=\text{avg}\left\Vert \frac{1}{n_{i}}\frac{\partial\mathcal{O}_{i}}{\partial\mathcal{D}_{\text{low}}}\right\Vert .
\]
This forces the gradients with respect to all outputs to be aligned simultaneously, regardless of their relative magnitude. The costs for $\mathcal{D}_{\text{mid}}$
and $\mathcal{D}_{\text{high}}$ are analogous. All training parameters
are the same as those in Subsection \ref{subsec:Results}. The cost function is 
\[
E=E_{0}+E_{1}+R_{\text{low}}^{I}+R_{\text{mid}}^{I}.
\]
Random directions for each input are cyclically shifted after each epoch. The results for the test set are shown in the first four columns of the Table \ref{tab:e vun 1}. The values for the training set and the remaining vulnerabilities are almost identical, so they were omitted.
\begin{table}
 \small
\caption{Test set error and average maximum vulnerability after first-order universal robust training\label{tab:e vun 1}}
\centering
\renewcommand{\arraystretch}{1.3}%
\begin{tabular}{|c|c|c|c|c|c|}
\hline 
\multicolumn{1}{|c|}{weights$\times$10\textsuperscript{6}} & 0.6 & 1.5 & 4 & 12 & 20* \tabularnewline
\hline 
\hline 
\multirow{1}{*}{$e$,\%} & 9.95 & 5.51 & 3.12 & 2.24 & 1.54\tabularnewline
\hline 
\multirow{1}{*}{$\bar{v}_{\text{max}}$,\%} & 21.1 & 17.3 & 13.3 & 11.6 & 7.50\tabularnewline
\hline 
\end{tabular}
\end{table}
The task is more demanding, as the networks must produce a specific output in a 1681-dimensional layer around the 6-dimensional manifold. Although accuracy has significantly decreased, it can be improved simultaneously with robustness by increasing the number of weights, adding more layers (while maintaining the same total number of weights), introducing more random directions, decreasing the batch size, or extending the number of epochs. To demonstrate the combined effect, we train another network with $L=2048$, two additional hidden layers (20 million weights in total), 84 batches, 2200 epochs, and a cost function that includes a high-frequency direction term
\[
E=E_{0}+E_{1}+R_{\text{low}}^{I}+R_{\text{mid}}^{I}+R_{\text{high}}^{I}.
\]
The result is shown in the last column of Table \ref{tab:e vun 1}, with the asterisk indicating the change in training conditions. An example of the network's gradient is shown in Fig. \ref{fig:g_robust}. Although robustness and accuracy can improve simultaneously for larger models, a trade-off still exists for each network, and tuning this trade-off may be beneficial. Specifically, adding one more epoch with a non-robust cost $E=E_{0}+E_{1}$ increases the accuracy by 3 to 4 times at the cost of a slight decrease in robustness. The results are shown in Table \ref{tab:e vun 1a}.
\begin{table}
 \small
\caption{Error and average vulnerabilities after first-order
universal robust training (trade-off adjusted)\label{tab:e vun 1a}}
\centering
\renewcommand{\arraystretch}{1.3}%
\begin{tabular}{|c|c|c|c|c|c|c|}
\hline 
\multicolumn{2}{|c|}{weights$\times$10\textsuperscript{6}} & 0.6 & 1.5 & 4 & 12 & 20* \tabularnewline
\hline 
\hline 
\multirow{2}{*}{$e$,\%} & training & 3.66 & 1.30 & 0.73 & 0.54 & 0.37\tabularnewline
\cline{2-7} \cline{3-7} \cline{4-7} \cline{5-7} \cline{6-7} \cline{7-7} 
 & test & 3.69 & 1.33 & 0.77 & 0.59 & 0.46\tabularnewline
\hline 
\hline 
\multirow{2}{*}{$\bar{v}_{\text{max}}$,\%} & training & 23.4 & 18.6 & 13.9 & 12.0 & 7.68\tabularnewline
\cline{2-7} \cline{3-7} \cline{4-7} \cline{5-7} \cline{6-7} \cline{7-7} 
 & test & 23.5 & 18.7 & 14.0 & 12.1 & 7.92\tabularnewline
\hline 
\hline 
\multirow{2}{*}{$\bar{v}_{\text{sen.}}$,\%} & training & 23.4 & 18.6 & 13.9 & 12.0 & 7.68\tabularnewline
\cline{2-7} \cline{3-7} \cline{4-7} \cline{5-7} \cline{6-7} \cline{7-7} 
 & test & 23.4 & 18.7 & 14.0 & 12.1 & 7.90\tabularnewline
\hline 
\hline 
\multirow{2}{*}{$\bar{v}_{\text{inv.}}$,\%} & training & 18.2 & 15.3 & 12.3 & 10.9 & 7.32\tabularnewline
\cline{2-7} \cline{3-7} \cline{4-7} \cline{5-7} \cline{6-7} \cline{7-7} 
 & test & 18.2 & 15.4 & 12.4 & 11.0 & 7.53\tabularnewline
\hline 
\end{tabular}
\end{table}
 
\subsubsection{Gaussian Augmentation}

As a baseline for robust training, we employ a simple data
augmentation \cite{zantedeschi_efficient_2017,gilmer_adversarial_2019}.
At the beginning of each epoch, Gaussian noise $\mathcal{N_{\sigma}}$ with variance
$\sigma^{2}$ and zero mean is added to clean images, so the problem is
\begin{equation}
N\left(\mathcal{I}\left(C\right)+\mathcal{N}_{\sigma}\right)=C.\label{eq:problem-noise}
\end{equation}
We start with a moderate amount of noise $\sigma=0.055$; the results are in Table \ref{tab:gaussian}.
\begin{table}
 \small
\caption{Error and average vulnerabilities for clean images after training 
with Gaussian augmentation, $\sigma=0.055$\label{tab:gaussian}}
\centering
\renewcommand{\arraystretch}{1.3}%
\begin{tabular}{|c|c|c|c|c|c|c|}
\hline 
\multicolumn{2}{|c|}{weights$\times$10\textsuperscript{6}} & 0.6 & 1.5 & 4 & 12 & 20*\tabularnewline
\hline 
\hline 
\multirow{2}{*}{$e$,\%} & training & 1.87 & 1.61 & 1.65 & 2.31 & 3.42\tabularnewline
\cline{2-7} \cline{3-7} \cline{4-7} \cline{5-7} \cline{6-7} \cline{7-7} 
 & test & 2.17 & 1.93 & 2.47 & 3.98 & 4.91\tabularnewline
\hline 
\hline 
\multirow{2}{*}{$\bar{v}_{\text{max}}$,\%} & training & 24.6 & 21.0 & 18.0 & 19.1 & 24.3\tabularnewline
\cline{2-7} \cline{3-7} \cline{4-7} \cline{5-7} \cline{6-7} \cline{7-7} 
 & test & 24.7 & 21.1 & 18.3 & 19.6 & 25.1\tabularnewline
\hline 
\hline 
\multirow{2}{*}{$\bar{v}_{\text{sen.}}$,\%} & training & 24.3 & 20.3 & 16.5 & 12.6 & 8.20\tabularnewline
\cline{2-7} \cline{3-7} \cline{4-7} \cline{5-7} \cline{6-7} \cline{7-7} 
 & test & 24.3 & 20.4 & 17.0 & 15.6 & 15.7\tabularnewline
\hline 
\hline 
\multirow{2}{*}{$\bar{v}_{\text{inv.}}$,\%} & training & 18.6 & 16.9 & 15.2 & 16.0 & 18.0\tabularnewline
\cline{2-7} \cline{3-7} \cline{4-7} \cline{5-7} \cline{6-7} \cline{7-7} 
 & test & 18.7 & 17.0 & 15.3 & 15.6 & 16.8\tabularnewline
\hline 
\end{tabular}
\end{table}
The network with 4 million weights demonstrates the lowest vulnerability,
while larger models overfit in several ways. First, pockets of invariance form around each training pattern, as evidenced by the overfitting of $\bar{v}_{\text{sen.}}$. This is expected, as condition (\ref{eq:problem-noise}) forces the network to be locally constant
while changing globally, and networks with lower capacity cannot fit such a target. Second, the networks overfit to the specific
amount of Gaussian noise \cite{kireev_effectiveness_2022}: although
the training error on noisy inputs (not shown) decreases slowly with
network size, the error on clean test samples, as well as for other
$\sigma$, increases.

\begin{figure}
\centering
\includegraphics[height=7.0cm, trim=0.3cm 0.25cm 0.2cm 0.2cm, clip]{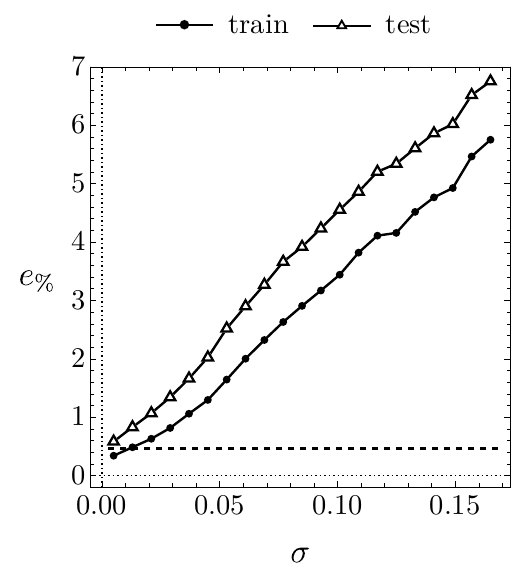}
\caption{Error for clean images after training with Gaussian augmentation. The dashed
line represents the test set error after universal robust training\label{fig:gaussian err}}
\end{figure}

\begin{figure}
\centering
\includegraphics[height=7.0cm, trim=0.20cm 0.25cm 0.18cm 0.2cm, clip]{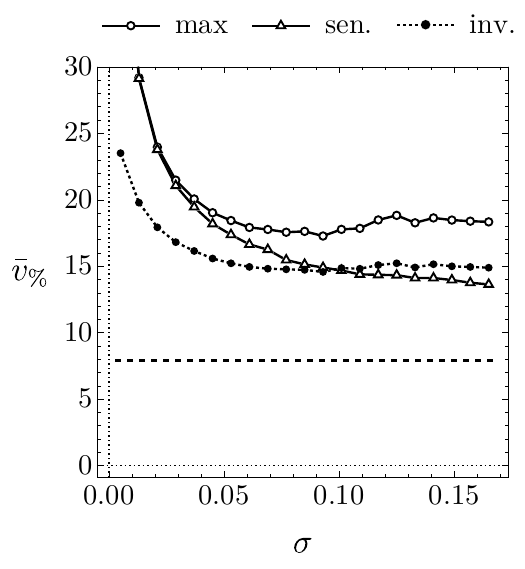} 
\caption{Average vulnerabilities for clean images after training with Gaussian augmentation. The dashed
lines represent the average maximum vulnerability
after universal robust training\label{fig:gaussian vun}}
\end{figure}

To examine how the amount of noise affects the training, we use the best-performing architecture with 4 million weights and train it with $\sigma\in\left[0.005,0.17\right]$. The results are shown in
Figs. \ref{fig:gaussian err} and \ref{fig:gaussian vun}.
Training and test errors increase steadily with $\sigma$. The vulnerability to
attacks exploiting sensitivity decreases monotonically. However, starting
at $\sigma=0.1$, invariance attacks become more effective and the overall
vulnerability begins to increase. This aligns with the reported trade-off between
different types of adversarial vulnerabilities \cite{tramer_fundamental_2020}.
The highest robustness is achieved for $\sigma=0.095$, and this network will be used for further comparisons. Its gradient is shown
in Fig. \ref{fig:g_gauss}. While it is indeed ``perceptually aligned''
\cite{santurkar_image_2019,kaur_are_2019} -- adding it to the original
image visibly resembles a cube motion, its quality is much lower than that achieved by the proposed universal robust training. It is clear that the results obtained with Gaussian augmentation depend on the balance between a contradictory cost function and the network's limited capabilities. Similar results can be obtained by minimizing
\[
E=E_{0}+c^{2}\sum_{\substack{\textrm{training}\\
\textrm{set}
}
}\left\Vert \partial N/\partial\mathcal{D}\right\Vert ^{2},
\]
which is a variant of Jacobian regularization \cite{drucker_improving_1992,jakubovitz_improving_2018,hoffman_robust_2019,chan_jacobian_2020}.
The key difference with the expression (\ref{eq:R1low}) is that
the derivative $\partial\mathcal{O}/\partial\mathcal{D}$ is discarded.
Although this derivative has a small absolute value, especially for $\mathcal{D}_{\text{high}}$, assuming it to be zero results in the same negative effects as with Gaussian augmentation. As $c^2$ increases, the values of $v_\text{sen.}$, $v_\text{inv.}$ and $v_\text{max}$ decrease until $v_\text{max}$ reaches a minimum of $18\%$. Beyond this point, invariance attacks become dominant, and the maximum vulnerability increases. Since this is worse than the best result achieved with Gaussian augmentation, we did not include detailed results. The visual quality of the network gradients is similar to that in Fig. \ref{fig:g_gauss}.

\section{Second order\label{sec:second-order}}
First-order robust training aligns the network gradients with the oracle only on the 6-dimensional manifold of clean images. Moving off this manifold (e.g., by adding noise) weakens the alignment and reduces robustness. This issue can be addressed by also aligning Hessians for clean images, which requires training all second derivatives.  Additionally, training second derivatives can further improve accuracy in the non-robust formulation. We begin by calculating second image derivatives.
\subsection{Image Derivatives}

\begin{figure}
\centering
\subfloat[]{\includegraphics[height=3.2cm]{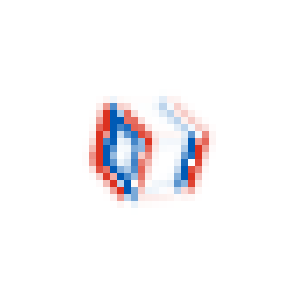}
}\includegraphics[height=2.8cm, trim=0.09cm 0.6cm 0.09cm 0.15cm, clip]{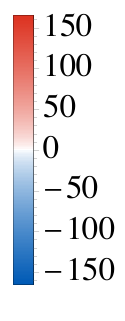}\subfloat[]{\includegraphics[height=3.2cm]{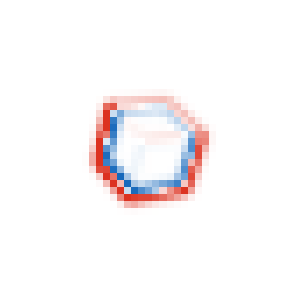}
}\includegraphics[height=2.8cm, trim=0.1cm 0.6cm 0.09cm 0.15cm, clip]{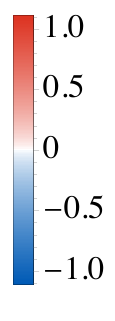}
\subfloat[]{\includegraphics[height=3.2cm]{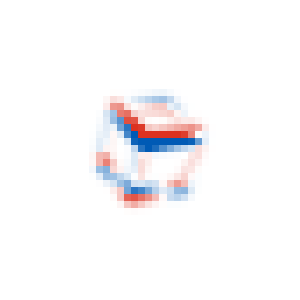}}\includegraphics[height=3.0cm, trim=0.2cm 0.3cm 0.0cm 0.16cm, clip]{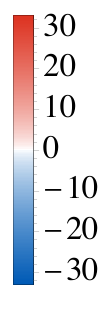}\subfloat[]{\includegraphics[height=3.2cm]{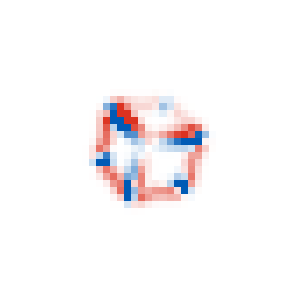}
}\includegraphics[height=3cm, trim=-0.04cm 0.3cm 0.09cm 0.15cm, clip]{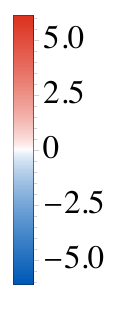}
\caption{Second-order image derivatives corresponding to pairs of infinitesimal
operators: $\text{Transl}.x\circ\text{Transl}.x$ (a), $\text{Transl}.z\circ\text{Transl}.z$ (b), $\text{Transl}.y\circ\text{Rot}.x$ (c) and $\text{Rot}.z\circ\text{Rot}.z$ (d)\label{fig:Second-derivatives}}
\end{figure}

The goal is to compute the second derivatives of the cube images with
respect to the parameters of the operator ($\text{\ref{eq:T}}$). This can be done by taking finite differences of first derivatives, evaluated for nearby cubes $\hat{T}\left(\pm\varepsilon\delta_{ik}\right)C$:
\begin{equation}
\frac{\partial^{2}\mathcal{I}}{\partial\nu_{i}\partial\nu_{j}}\simeq\frac{1}{2\varepsilon}\left(\frac{\partial\mathcal{I}\left(\hat{T}\left(\varepsilon\delta_{ik}\right)C\right)}{\partial\nu_{j}}-\frac{\partial\mathcal{I}\left(\hat{T}\left(-\varepsilon\delta_{ik}\right)C\right)}{\partial\nu_{j}}\right)\label{eq:d2fd}
\end{equation}
Note that the operator $\hat{T}$ does not commute with a similar operator from the definition of first-order derivative $\partial \mathcal{I} / \partial \nu $. For this expression to yield the second derivatives of the original operator ($\text{\ref{eq:T}}$), we require $i\leq j$. The case $i>j$ corresponds to another version of ($\text{\ref{eq:T}}$) with a different order of operations. In total, there are 21 second derivatives, which can be computed with relative accuracy of $10^{-6}$ by setting $\varepsilon=5\cdot10^{-6}$. Examples are shown in Fig. \ref{fig:Second-derivatives}. The second derivatives of $C$ are obtained by differentiating (\ref{eq:T}) twice
and substituting $\nu_{i}=0$. Most of these derivatives are zero, except for the six corresponding to rotations ($i,j\leq3$).
 Denoting the image Hessian
\[
\mathcal{H}_{ij}=\frac{\partial^{2}\mathcal{I}}{\partial\nu_{i}\partial\nu_{j}},
\]
the second-order expansions of the cube's vertices and image are
\begin{equation}
\hat{T}\left(\nu_{i}\right)C=C+\frac{\partial C}{\partial\nu_{i}}\nu_{i}+\frac{1}{2}\frac{\partial^{2}C}{\partial\nu_{i}\partial\nu_{j}}\nu_{i}\nu_{j}+o\left(\nu^{2}\right),
\end{equation}
\begin{equation}
\mathcal{I}\left(\hat{T}\left(\nu_{i}\right)C\right)=\mathcal{I}\left(C\right)+\frac{\partial\mathcal{I}}{\partial\nu_{i}}\nu_{i}+\frac{1}{2}\mathcal{H}_{ij}\nu_{i}\nu_{j}+o\left(\nu^{2}\right).\label{eq:second order series}
\end{equation}

\subsection{Improving Accuracy\label{subsec:accuracy-2}}

The procedure is very similar to that described in Subsection \ref{subsec:accuracy-1}.
Second-order derivatives are now included in the forward pass
\[
\mathcal{I},\frac{\partial\mathcal{I}}{\partial\nu_{i}},\frac{\partial^{2}\mathcal{I}}{\partial\nu_{i}\partial\nu_{j}}\overset{\substack{\textrm{forward}\\
\textrm{pass}
}
}{\longrightarrow}N,\frac{\partial N}{\partial\nu_{i}},\frac{\partial^{2}N}{\partial\nu_{i}\partial\nu_{j}},
\]
and the second-order cost function is
\[
E_{2}=\sum_{\substack{\textrm{training}\\
\textrm{set}
}
}\left[\sum_{i=1}^{6}\sum_{j=i}^{6}\frac{1}{n_{ij}^{2}}\left\Vert \frac{\partial^{2}N}{\partial\nu_{i}\partial\nu_{j}}-\frac{\partial^{2}\mathcal{O}}{\partial\nu_{i}\partial\nu_{j}}\right\Vert ^{2}\right].
\]
Normalization factors $n_{ij}$ require more consideration. For the 6 rotational derivatives, the target is non-zero, so we can use
\[
n_{ij}=\text{avg}\left\Vert \frac{\partial^{2}\mathcal{O}}{\partial\nu_{i}\partial\nu_{j}}\right\Vert ,j,i\leq3.
\]
These $n_{ij}$ are of the order of 1, and the average magnitude of the corresponding
input derivatives is
\[
\underset{j,i\leq3}{\textrm{avg}}\left\Vert \frac{\partial\mathcal{I}}{\partial\nu_{i}\partial\nu_{j}}\right\Vert \simeq40.
\]
Since the output scales linearly with the input, the normalization factors for the remaining derivatives with zero targets can be estimated as
\[
n_{ij}=40/\textrm{avg}\left\Vert \frac{\partial\mathcal{I}}{\partial\nu_{i}\partial\nu_{j}}\right\Vert ,i>3.
\]
The cost function is $E=E_{0}+E_{1}+E_{2}$.
The initial training lasts for 1950 epochs, after which the learning rate is reduced to $l=10^{-4}$
for 25 epochs. Following the approach of gradual exclusion of higher derivatives
described in \cite{avrutskiy_enhancing_2020}, the second derivatives are dropped and 25 additional epochs are run with the learning rate $l=10^{-5}$. The results are shown
in Table \ref{tab:accuracy-d2}. The accuracy increases by a factor of 25 compared to conventional training and by a factor of 4 compared to first-order training, once again closely aligning with results from low-dimensional function approximation.
\begin{table}
 \small
\caption{
Error in vertex reconstruction after second order training\label{tab:accuracy-d2}}
\centering
\renewcommand{\arraystretch}{1.3}%
\begin{tabular}{|c|c|c|c|c|c|}
\hline 
\multicolumn{2}{|c|}{weights$\times$10\textsuperscript{6}} & 0.6 & 1.5 & 4 & 12\tabularnewline
\hline
\hline
\multirow{2}{*}{$e$,\%} & training & 0.111 & 0.044 & 0.017 & 0.013\tabularnewline
\cline{2-6} \cline{3-6} \cline{4-6} \cline{5-6} \cline{6-6} 
 & test & 0.112 & 0.046 & 0.021 & 0.018\tabularnewline
\hline 
\end{tabular}
\end{table}

\subsection{Oracle Expansion\label{subsec:Oracle's-second-derivatives}}

Hessian alignment can be achieved by minimizing the deviation between the second derivatives of the network and the oracle. The second derivatives of the network are
computed via forward pass. The second derivative of the oracle is \begin{equation}
\begin{split}
\frac{\partial^{2}\mathcal{O}}{\partial\mathcal{D}^{2}}=\lim_{\varepsilon\rightarrow0}\frac{1}{\varepsilon^{2}}\Bigl(\mathcal{O}\left(\mathcal{I}\left(C\right)+\varepsilon\mathcal{D}\right)+\mathcal{O}&\left(\mathcal{I}\left(C\right)-\varepsilon\mathcal{D}\right)-\\&-2\mathcal{O}\left(\mathcal{I}\left(C\right)\right)\Bigr).\label{eq:oracle j2}
\end{split}
\end{equation}
As before, the output of the oracle is expressed as the transformation of the initial cube
\[
\mathcal{O}\left(\mathcal{I}\left(C\right)+\varepsilon\mathcal{D}\right)=\hat{T}\left(\nu_{i}\left(\varepsilon\right)\right)C,
\]
and the second-order expansion for its parameters is
\begin{equation}
\nu_{i}\left(\varepsilon\right)=\varepsilon\frac{d\nu_{i}}{d\varepsilon}+\frac{\varepsilon^{2}}{2}\frac{d^{2}\nu_{i}}{d\varepsilon^{2}}+o\left(\varepsilon^{2}\right).\label{eq:nu so}
\end{equation}
The first term has already been calculated. To obtain the next term
$d^{2}\nu_{i}/d\varepsilon^{2}$, we must solve a minimization
problem
\begin{equation}
\min_{d^{2}\nu_{i}/d\varepsilon^{2}}\left\Vert \mathcal{I}\left(\hat{T}\left(\varepsilon\frac{d\nu_{i}}{d\varepsilon}+\frac{\varepsilon^{2}}{2}\frac{d^{2}\nu_{i}}{d\varepsilon^{2}}\right)C\right)-\mathcal{I}\left(C\right)-\varepsilon\mathcal{D}\right\Vert ^{2}.\label{eq:d2v/de2 init}
\end{equation}
Substituting (\ref{eq:nu so}) into the second-order expansion of
the images ($\text{\ref{eq:second order series}}$) yields 
\begin{align}
\begin{split}
&\mathcal{I}\left(\hat{T}\left(\varepsilon\frac{d\nu_{i}}{d\varepsilon}+\frac{\varepsilon^{2}}{2}\frac{d^{2}\nu_{i}}{d\varepsilon^{2}}\right)C\right)=\mathcal{I}\left(C\right)+\varepsilon\frac{\partial\mathcal{I}}{\partial\nu_{i}}\frac{d\nu_{i}}{d\varepsilon}+\\&+\frac{\varepsilon^{2}}{2}\left(\frac{\partial\mathcal{I}}{\partial\nu_{i}}\frac{d^{2}\nu_{i}}{d\varepsilon^{2}}+\mathcal{H}_{ij}\frac{d\nu_{i}}{d\varepsilon}\frac{d\nu_{j}}{d\varepsilon}\right)+\frac{\varepsilon^{3}}{2}\mathcal{H}_{ij}\frac{d\nu_{i}}{d\varepsilon}\frac{d^{2}\nu_{j}}{d\varepsilon^{2}}.\label{eq:2nd exp}
\end{split}
\end{align}
The last summand may seem out of order, but the third-order expansion
terms proportional to $\varepsilon^{3}$ do not depend on $d^{2}\nu_{i}/d\varepsilon^{2}$,
meaning that for our purposes, this relationship is $o\left(\varepsilon^{3}\right)$
accurate. After substituting it into ($\text{\ref{eq:d2v/de2 init}}$)
and expanding the squared norm, terms proportional to $\varepsilon^{3}$
cancel out when considering the first-order relation ($\text{\ref{eq:first directional}}$).
Therefore, we only need to consider terms proportional to $\varepsilon^{4}$:
\begin{align*}
\min_{d^{2}\nu_{i}/d\varepsilon^{2}}\left[\left(\frac{\partial\mathcal{I}}{\partial\nu_{i}}\frac{d\nu_{i}}{d\varepsilon}-\mathcal{D}\right)\cdot\mathcal{H}_{ij}\frac{d\nu_{i}}{d\varepsilon}\frac{d^{2}\nu_{j}}{d\varepsilon^{2}}\right.+\\\left.+\frac{1}{4}\left\Vert \frac{\partial\mathcal{I}}{\partial\nu_{i}}\frac{d^{2}\nu_{i}}{d\varepsilon^{2}}+\mathcal{H}_{ij}\frac{d\nu_{i}}{d\varepsilon}\frac{d\nu_{j}}{d\varepsilon}\right\Vert ^{2}\right].
\end{align*}
By expanding the squared norm and taking derivatives with respect
to $d^{2}\nu_{i}/d\varepsilon^{2}$ we get a system of 6 equations
\begin{equation}
\begin{split}
\Gamma_{ij}\frac{d^{2}\nu_{j}}{d\varepsilon^{2}}&=2\left(\mathcal{D}\cdot\mathcal{H}_{ik}\right)\frac{d\nu_{k}}{d\varepsilon}-2\left(\mathcal{H}_{ki}\cdot\frac{\partial\mathcal{I}}{\partial\nu_{p}}\right)\frac{d\nu_{p}}{d\varepsilon}\frac{d\nu_{k}}{d\varepsilon}-\\&-\left(\mathcal{H}_{kp}\cdot\frac{\partial\mathcal{I}}{\partial\nu_{i}}\right)\frac{d\nu_{p}}{d\varepsilon}\frac{d\nu_{k}}{d\varepsilon}.\label{eq:s.directional-1}
\end{split}
\end{equation}
Similar to ($\text{\ref{eq:first directional}}$), the solution can be expressed using $\Gamma^{-1}$. The output of the oracle is
\begin{align*}
\mathcal{O}\left(\mathcal{I}\left(C\right)+\varepsilon \mathcal D\right)&=C+\varepsilon\frac{\partial C}{\partial\nu_{i}}\frac{d\nu_{i}}{d\varepsilon}+\\&+\frac{\varepsilon^{2}}{2}\left(\frac{\partial C}{\partial\nu_{i}}\frac{\partial^{2}\nu_{i}}{d\varepsilon^{2}}+\frac{\partial^{2}C}{\partial\nu_{i}\partial\nu_{j}}\frac{d\nu_{j}}{d\varepsilon}\frac{d\nu_{i}}{d\varepsilon}\right)+o\left(\varepsilon^{2}\right),
\end{align*}
by substituting it into ($\text{\ref{eq:oracle j2}}$) we get
\[
\frac{\partial^{2}\mathcal{O}}{\partial\mathcal{D}^{2}}=\frac{\partial C}{\partial\nu_{i}}\frac{\partial^{2}\nu_{i}}{d\varepsilon^{2}}+\frac{\partial^{2}C}{\partial\nu_{i}\partial\nu_{j}}\frac{d\nu_{i}}{d\varepsilon}\frac{d\nu_{j}}{d\varepsilon},
\]
where $d\nu_{j}/d\varepsilon$ is the solution of ($\text{\ref{eq:first directional}}$)
and $\partial^{2}\nu_{i}/d\varepsilon^{2}$ is the solution of ($\text{\ref{eq:s.directional-1}}$).

\subsection{Universal Robust Training\label{subsec:Robust-training-2}}

The expression and normalization for second-order universal robust training are similar those for first-order training
\[
R_{\text{low}}^{I\!I}=\frac{1}{c^{2}}\sum_{\substack{\textrm{training}\\
\textrm{set}
}
}\left[\sum_{i=1}^{9}\frac{1}{n_{i}^{2}}\left(\frac{\partial^{2}N_{i}}{\partial\mathcal{D}_{\text{low}}^{2}}-\frac{\partial^{2}\mathcal{O}_{i}}{\partial\mathcal{D}_{\text{low}}^{2}}\right)^{2}\right],
\]
\[
n_{i}=P_{95}\left[\frac{\partial^{2}\mathcal{O}_{i}}{\partial\mathcal{D}_{\text{low}}^{2}}\right]-P_{5}\left[\frac{\partial^{2}\mathcal{O}_{i}}{\partial\mathcal{D}_{\text{low}}^{2}}\right],
\]
\[
c=\text{avg}\left\Vert \frac{1}{n_{i}}\frac{\partial^{2}\mathcal{O}_{i}}{\partial\mathcal{D}_{\text{low}}^{2}}\right\Vert.
\]
All training parameters are the same as those in Subsection~\ref{subsec:Robust-training}. The first four networks are trained using
\[
E=E_{0}+E_{1}+R_{\text{low}}^{I}+R_{\text{mid}}^{I}+R_{\text{low}}^{I\!I}+R_{\text{mid}}^{I\!I}.
\]
The network with 20 million weights is trained with two additional terms, $R_{\text{high}}^{I}+R_{\text{high}}^{I\!I}$. The results for the clean images from the test set are shown in Table \ref{tab:r2}. The values for the training set and the remaining vulnerabilities are nearly identical, so they have been omitted.
\begin{table}
 \small
\caption{Test error and average maximum vulnerability after second-order universal robust training\label{tab:r2}}
\centering
\renewcommand{\arraystretch}{1.3}%
\begin{tabular}{|c|c|c|c|c|c|}
\hline 
\multicolumn{1}{|c|}{weights$\times$10\textsuperscript{6}} & 0.6 & 1.5 & 4 & 12 & 20* \tabularnewline
\hline 
\hline 
\multirow{1}{*}{$e$,\%} & 22.9 & 14.9 & 8.42 & 6.21 & 4.71\tabularnewline
\hline 
\multirow{1}{*}{$\bar{v}_{\text{max}}$,\%} & 20.4 & 17.0 & 13.0 & 11.2 & 7.20\tabularnewline
\hline 
\end{tabular}
\end{table}
As before, by adding one extra epoch with the cost function $E=E_{0}+E_{1}$,
we can significantly increase the accuracy at the cost of a small decrease in robustness. The results are shown in Table \ref{tab:r2-e}. Apart from a slight decrease in accuracy and a larger gap between networks, the overall results remain very similar to those from first-order training.
\begin{table}
\small
\caption{Error and average vulnerabilities after second-order
robust training, trade-off adjusted\label{tab:r2-e}}
\centering
\renewcommand{\arraystretch}{1.3}%
\begin{tabular}{|c|c|c|c|c|c|c|}
\hline 
\multicolumn{2}{|c|}{weights$\times$10\textsuperscript{6}} & 0.6 & 1.5 & 4 & 12 & 20* \tabularnewline
\hline 
\hline 
\multirow{2}{*}{$e$,\%} & training & 12.7 & 5.59 & 1.90 & 1.20 & 0.69\tabularnewline
\cline{2-7} \cline{3-7} \cline{4-7} \cline{5-7} \cline{6-7} \cline{7-7} 
 & test & 12.7 & 5.62 & 1.92 & 1.23 & 0.73\tabularnewline
\hline 
\hline 
\multirow{2}{*}{$\bar{v}_{\text{max}}$,\%} & training & 22.4 & 18.8 & 14.1 & 12.0 & 7.50\tabularnewline
\cline{2-7} \cline{3-7} \cline{4-7} \cline{5-7} \cline{6-7} \cline{7-7} 
 & test & 22.4 & 18.9 & 14.2 & 12.1 & 7.58\tabularnewline
\hline 
\hline 
\multirow{2}{*}{$\bar{v}_{\text{sen.}}$,\%} & training & 20.7 & 18.4 & 14.1 & 12.0 & 7.48\tabularnewline
\cline{2-7} \cline{3-7} \cline{4-7} \cline{5-7} \cline{6-7} \cline{7-7} 
 & test & 20.7 & 18.5 & 14.1 & 12.1 & 7.55\tabularnewline
\hline 
\hline 
\multirow{2}{*}{$\bar{v}_{\text{inv.}}$,\%} & training & 19.6 & 16.3 & 12.7 & 11.0 & 7.22\tabularnewline
\cline{2-7} \cline{3-7} \cline{4-7} \cline{5-7} \cline{6-7} \cline{7-7} 
 & test & 19.6 & 16.4 & 12.7 & 11.1 & 7.29\tabularnewline
\hline
\end{tabular}
\captionsetup[subfloat]{captionskip=0pt}
\end{table}

\begin{figure}
\centering
\includegraphics[height=7cm, trim=0.1cm 0.6cm 0.0cm 0.2cm, clip]{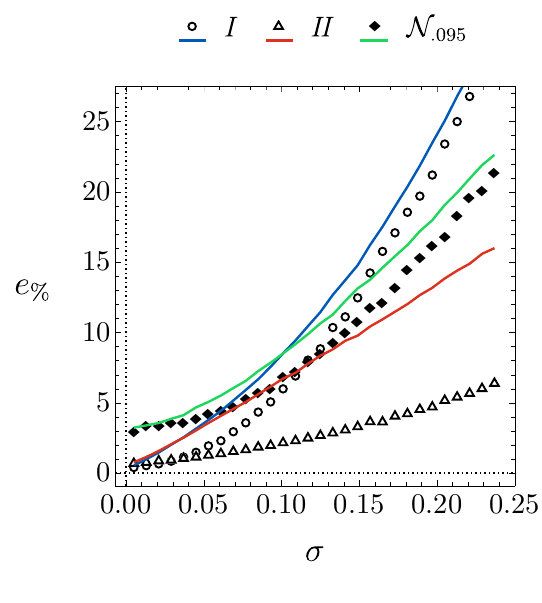}
\caption{Error in vertex reconstruction under Gaussian noise after first-order ($I$), second-order ($I\!I$) universal robust training and the best result from Gaussian augmentation ($\mathcal{N}_{.095}$). Colored lines represent results without correction (\ref{eq:corr})\label{fig:cor-res-acc}}
\end{figure}

The purpose of the second-order training, however, was to extend the
robustness further into the 1681-dimensional image space. To assess whether we succeeded, we evaluate the accuracy and average maximum vulnerability under Gaussian noise of various magnitudes for images in the test set. Since adding noise may alter the oracle's output, the minimization problem (\ref{eq:extended problem}) must be reassessed before accuracy or robustness can be estimated. Fortunately, the Gaussian noise produces relatively small changes that can be accurately captured
by the second-order Taylor expansion (\ref{eq:oracle expansion}) up to $\sigma=0.25$. Substituting
$\mathcal{D}=\mathcal{N}_{\sigma}$ and $\varepsilon=1$, we obtain
\begin{equation}
\mathcal{O}\left(\mathcal{I}+\mathcal{N}_{\sigma}\right)\simeq\mathcal{O}\left(\mathcal{I}\right)+\frac{\partial\mathcal{O}}{\partial\mathcal{N}_{\sigma}}+\frac{1}{2}\frac{\partial^{2}\mathcal{O}}{\partial\mathcal{N}_{\sigma}^{2}}.\label{eq:corr}
\end{equation}
The first-order term dominates the adjustment, while the second-order
contributes up to 1/20 of the correction, and the remaining discrepancy is less than $0.05\%$. After the new vertices are obtained, image derivatives are generated
and the maximum vulnerability is determined by solving (\ref{eq:max vun}). Fig. \ref{fig:cor-res-acc} shows the results after first-order ($I$) and second-order ($I\!I$) universal robust training for the network with 20 million weights,
compared to the best result of Gaussian augmentation ($\mathcal{N}_{.095}$). The
results without correction (\ref{eq:corr}), corresponding to the output invariance assumption, are shown in color. Two details are worth mentioning: first, the noise-induced geometry correction improves the outcome even for Gaussian augmentation. Second, even under the invariance assumption, the accuracy of $I\!I$ is consistently higher than $\mathcal{N}_{.095}$. Although, for heavier noise, the accuracy of $I$ drops below that of $\mathcal{N}_{.095}$, the results of $I\!I$ remain unmatched.
\begin{figure}
\centering
\includegraphics[height=7.00cm, trim=0.1cm 0.6cm 0.0cm 0.2cm, clip]{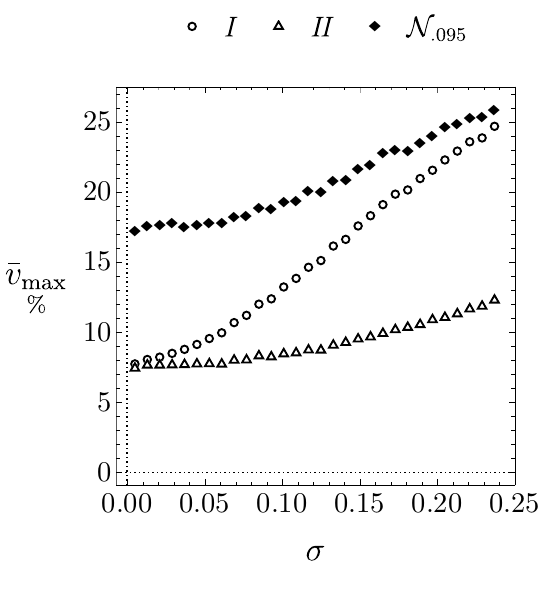}

\caption{Average maximum vulnerability under Gaussian noise
after first-order ($I$), second-order ($I\!I$)
universal robust training and the best result from Gaussian augmentation ($\mathcal{N}_{.095}$)\label{fig:cor-res-vun}}
\end{figure}
Fig. \ref{fig:cor-res-vun} shows the average maximum vulnerability. The
results without correction (\ref{eq:corr}) differ by no more than 1\%, so they are not shown. Network $I$ gradually loses its robustness, but as expected, the second-order results remain much more stable. The effect of tradeoff adjustment is uniform across $\sigma$ for both plots. It is worth noting that for $\sigma=0.25$ the distance between the original and noisy image is approximately 10, roughly matching the original manifold's diameter. Thus, training derivatives only on the manifold can significantly improve error in a substantial region of the full image space.

\section{Related Work}

\subsection{Robust Training}

Since standard adversarial training against sensitivity-based attacks
typically comes at the cost of accuracy \cite{tsipras_robustness_2018,su_is_2018,dobriban_provable_2023},
several works have focused on improving this trade-off \cite{tanay_boundary_2016,lopes_improving_2019,arani_adversarial_2020,hu_triple_2020,yang_closer_2020}.
Works \cite{nakkiran_adversarial_2019,stutz_disentangling_2019} presented
the evidence that this trade-off can be avoided entirely. The discovery
of invariance-based attacks introduced a new robustness metric,
which turned out to involve a trade-off with the previous one \cite{tramer_fundamental_2020,singla_shift_2021}.
The trade-off between two types of robustness and accuracy was addressed through Pareto optimization \cite{sun_ke_pareto_2023} as well as other
balancing approaches \cite{chen_balanced_2022,rauter_effect_2023}. Work
\cite{yang_one_2022} suggested weakening the invariance radius for
each input individually. In this paper, we demonstrate how both types of robustness can be achieved simultaneously with high accuracy, with network capacity appearing to be the only limiting factor.

\subsection{Training Derivatives\label{subsec:td-ref}}

There are two distinct approaches to training derivatives. The first and most common, uses them as a regularization for the initial
problem \cite{belkin_manifold_2006}, aiming to enforce $D\left(N\right)=0$
for a specific set of derivatives. This approach was introduced in \cite{drucker_improving_1992,goos_transformation_1998}
for classifiers, in extended in \cite{rifai_contractive_2011,rifai_higher_2011}
for autoencoders, and further developed in \cite{roth_stabilizing_2017,qi_global_2018,lecouat_semi-supervised_2018}
for generative adversarial networks. This method has also been applied to
adversarial training \cite{miyato_virtual_2018,ross_improving_2018,lecouat_semi-supervised_2018,qi_global_2018,jakubovitz_improving_2018,hoffman_robust_2019,chan_jacobian_2020}.
As noted earlier, the condition $D\left(N\right)=0$ can conflict with non-zero oracle gradients, resulting in a trade-off between different types of vulnerability. Moreover, this approach does not allow for any increase in the accuracy of the original problem.
The second, less common approach involves enforcing the condition $D\left(N\right)=D\left(\text{Target}\right)$, in addition to the objective $N=\text{Target}$.
This idea has been implemented in \cite{cardaliaguet_approximation_1992,basson_approximation_1999,flake_differentiating_1999,he_multilayer_2000,pukrittayakamee_practical_2011,yuan2022sobolev},
achieving moderate success. The author's paper \cite{avrutskiy_enhancing_2020}
demonstrated how this method can significantly improve accuracy in low dimensional cases. The present work achieves comparable improvements in solving an illustrative image analysis problem.

Another notable field related to training derivatives is the solution of differential equations using neural networks \cite{meade_jr_numerical_1994,lagaris_artificial_1997,lagaris_artificial_1998,budkina_neural_2016,tarkhov_semi-empirical_2019,kumar_multilayer_2011}.
In this approach, the network takes a vector of independent variables as input and outputs the corresponding solution value. $D$ represents the differential operator of the equation. The weights are trained to minimize the residual $D\left(N\right)$, while satisfying boundary and initial conditions. This flexible process allows for solving underdetermined, overdetermined \cite{karniadakis_physics-informed_2021}
and inverse problems \cite{gorbachenko_neural_2016,kryzhanovsky_solving_2022}.

\subsection{Tangent Plane}

The results of Section \ref{sec:first-order} are based on first-order analysis, where the data manifold is locally approximated by a tangent plane. We will reference studies that have constructed and utilized tangent planes for network training; however, all these studies adopt the derivatives-as-regularization approach, as outlined in Subsection~\ref{subsec:td-ref}. In \cite{goos_transformation_1998}, a classifier for handwritten digits
(MNIST) is considered. The tangent plane for each image is constructed
as a linear space of infinitesimal rotations, translations, and deformations.
A cost function enforces the invariance of the classifier under tangential
perturbations. In \cite{rifai_manifold_2011}, classifiers for MNIST and color images
from 10 different classes (CIFAR-10) are trained to remain invariant
under perturbations restricted to the tangent plane. The plane is
constructed using a contractive autoencoder \cite{rifai_contractive_2011}. In \cite{kumar_semi-supervised_2017} the classification task is solved
for CIFAR-10 and color street numbers (SVHN). An additional cost function
enforces the invariance with respect to tangential perturbations.
The plane is constructed using the generator part of the Generative
Adversarial Network (GAN). In work \cite{yu_tangent-normal_2019} classifiers for SVHN, CIFAR-10,
and grayscale fashion items in 10 categories (FashionMNIST) are considered.
The tangent plane is constructed using a variational autoencoder and localized
GAN \cite{qi_global_2018}. Tangential perturbations are regularized
through virtual adversarial training. Additional regularization with
respect to orthogonal perturbations is used to enforce robustness.

\subsection{Tracking Problems}

Although the goal of object tracking is to determine a bounding
box, several approaches achieve this through partial reconstruction of the tracked
object. This reconstruction is typically performed by minimizing a norm with respect to the parameters of a low-dimensional representation, similar to the oracle definition used in this paper. Low-dimensional
representations can be obtained using neural networks \cite{wang_learning_2013,jin_tracking_2013},
principal component analysis \cite{black_eigentracking_1998,mei_robust_2009,kwon_visual_2010}
or dictionary learning \cite{wang_online_2013}. While methods that do not rely on deep neural networks generally exhibit higher robustness, this paper focuses on training a robust deep network.

\section{Conclusion}

This study demonstrated how training derivatives can significantly enhance network accuracy for both noiseless and noisy images. We also showed that the fundamental trade-off between accuracy and robustness against sensitivity/invariance attacks can be eliminated, with all metrics improving by a factor of 2 to 4 compared to conventional robust training methods. Future research will focus on applying this method to practical problems where the data manifold is accessible, - i.e., its elements can be efficiently generated from a set of parameters. One such example is the phase retrieval problem, which involves recovering a function from the magnitude of its Fourier transform, particularly in Bragg Coherent Diffraction Imaging (BCDI). Neural networks have proven highly effective \cite{wang_use_2024} in this area. BCDI reconstructs the shape of a microcrystal from its 3D diffraction pattern, which is often quite noisy. Since the shapes can be parametrized \cite{wu_complex_2021}, the presented method can be applied directly. Another potential application is denoising algorithm based on Formula \ref{eq:corr}. If the manifold can be represented by a sufficiently smooth autoencoder, the oracle can be constructed and used to train a network that determines the optimal parameters of the latent representation with greater accuracy, as demonstrated in our illustrative example.

\section*{Acknowledgments}
The author would like to thank E.A. Dorotheyev and K.A. Rybakov for
fruitful discussions, and expresses sincere gratitude to I.A. Avrutskaya
and I.V. Avrutskiy, without whom this work would be impossible.
\bibliographystyle{IEEEtran}
\bibliography{bibgr}

\begin{IEEEbiography}[{\includegraphics[width=1.05in,height=1.5in,clip,keepaspectratio]{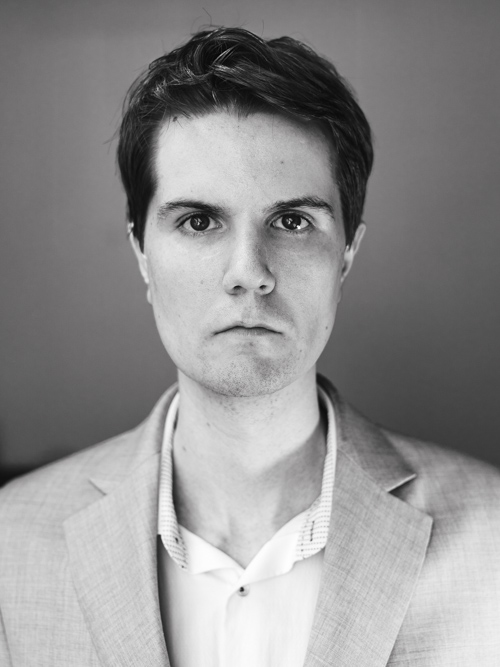}}]{Vsevolod I. Avrutskiy}
Received a Ph.D. in Machine Learning for Mathematical Physics from Moscow Institute of Physics and Technology (MIPT) in 2021. Worked at MIPT as an instructor in the Department of Theoretical Physics, Department of Computer Modelling, and as a researcher in the Laboratory of Theoretical Attosecond Physics. Current research interests include training neural network derivatives, shape optimization, and solving partial differential equations using neural networks.\end{IEEEbiography}

\clearpage
\appendices

\section{Image Generation}
\label{Image-gen}
The perspective image of a cube is created by projecting
its faces onto the image plane of a camera, which is defined by its position, orientation and focal length $\mathsf{f}$. The projection
$\left(u,v\right)$ of a single point $\left(x,y,z\right)$ can be obtained
through a linear transformation of vectors in homogeneous coordinates \cite{bloomenthal_homogeneous_1994}
\begin{equation}
\left(\begin{array}{c}
u\\
v\\
1
\end{array}\right)=\mathbf{P}\left(\begin{array}{c}
x\\
y\\
z\\
1
\end{array}\right).\label{eq:projection}
\end{equation}
The left-hand side implies that the values of $u$ and $v$ are obtained by normalizing the third component of the vector. If the image plane is aligned with the $\left(x,y\right)$ plane and the camera is pointing in the $z$-direction, the matrix $\mathbf{P}$ is
\[
\mathbf{P}=\left(\begin{array}{ccc}
\mathsf{f} & 0 & \mathsf{c_{x}}\\
0 & \mathsf{f} & \mathsf{c_{y}}\\
0 & 0 & 1
\end{array}\right)\left(\begin{array}{cccc}
R_{11} & R_{12} & R_{13} & t_{x}\\
R_{21} & R_{22} & R_{23} & t_{y}\\
R_{31} & R_{32} & R_{33} & t_{z}
\end{array}\right),
\]
where $\left(t_{x},t_{y},t_{z}\right)$ is the position of the camera, and $\left(\mathsf{c_{x}},\mathsf{c_{y}}\right)$ is origin of the image plane in $(x,y)$ coordinates. A 3 by 3 matrix $R_{ij}$ rotates the camera.

After projection, six faces of the cube produce up to three visible
quadrilaterals $Q$ in the $\left(u,v\right)$ plane
\begin{equation}
Q=\left\{ \left(u_{1},v_{1}\right),\left(u_{2},v_{2}\right),\left(u_{3},v_{3}\right),\left(u_{4},v_{4}\right)\right\} ,\label{eq:Q}
\end{equation}
which are colored according to the corresponding cube faces, with intensities ranging from $1/6$ to $1$; the empty space is black. Thus, intensity $I\left(u,v\right)$ is assigned to each point on the image plane. However, its values are not smooth functions of the cube's degrees of freedom. Additionally, probing this intensity on a grid produces an unnatural-looking image with sharp and irregular edges. To solve both issues, we implement an anti-aliasing procedure:
\begin{equation}
\bar{I}\left(u,v\right)=\frac{1}{\sigma\sqrt{2\pi}}\iint \! I\left(u',v'\right)e^{-\frac{1}{2\sigma^{2}}\left[\left(u-u'\right)^{2}+\left(v-v'\right)^{2}\right]}du'dv'.
\end{equation}
Since intensity is zero everywhere except for non-overlapping
quadrilaterals, this integral can be split
\[
\bar{I}\left(u,v\right)=\sum_{\text{visible }Q}\bar{I}\left(u,v,Q\right),
\]
and each visible quadrilateral contributes
\begin{align}
\bar{I}\left(u,v,Q\right) & =\frac{I\left(Q\right)}{\sigma\sqrt{2\pi}}\underset{Q}{\iint}e^{-\frac{1}{2\sigma^{2}}\left[\left(u-u'\right)^{2}+\left(v-v'\right)^{2}\right]}du'dv'\equiv\nonumber \\
 & \equiv I \left(Q\right)\underset{Q}{\iint}K\left(u,v,u',v'\right)du'dv',\label{eq:pixel}
\end{align}
where, $I \left( Q \right)$ is the intensity of the cube's face. 41 by 41 images are produced by sampling a square region $[-1,1]^2$ of the image plane using a uniform grid with step size $0.05$
\[ 
\mathcal{I}\left(C\right)=\left\{ \bar{I}\left(u,v\right)\mid u,v\in\left\{ -1,-0.95,\dots,1\right\} \right\}.
\]
The convolution parameter $\sigma=0.03$ was chosen to produce reasonably smooth images. The effective image resolution, based on the 50\% contrast loss in the Modulation Transfer Function (MTF) \cite{williams_introduction_2002}, is 28 by 28. As for the other parameters, the range of cube positions should ensure its projection remains within the image area. Additionally, the camera parameters and cube size should produce a reasonable amount of perspective distortion. We choose $\mathsf{f}=5$, $t_{x}=t_{y}=0$, $t_{z}=-5$, no shift ($\mathsf{c_{x}}=\mathsf{c_{y}}=0$) and no rotation ($R_{ij}=\delta_{ij}$).
The size of the cube is $0.4$. The coordinates of its center of mass can vary within the range $\left[-0.52,0.52\right]$. This results in a visible
width of 11 pixels at the closest position, 10 at the middle, and
9 at the farthest.

\section{Image Derivatives}
\label{Derivatives-of-images}
Even though first derivatives can be computed via finite differences, the rendering algorithm described in Appendix~\ref{Image-gen} allows for analytical expressions, significantly simplifying the computations. Upon applying the infinitesimal operators $\partial \hat{T}/\partial \nu_{i}$, cube's vertices acquire velocities that can be obtained by differentiating ($\text{\ref{eq:T}}$)
with respect to $\nu_{i}$, and setting $\nu_{i}=0$. The velocities of their image plane projections $\left(u,v\right)$ can be found from ($\text{\ref{eq:projection}}$)
by assuming $x=x\left(\nu_{i}\right)$, $y=y\left(\nu_{i}\right)$,
$z=z\left(\nu_{i}\right)$ and similarly evaluating derivatives with respect to
$\nu_{i}$ at $\nu_{i}=0$. For the chosen camera parameters
\[
\frac{\partial u}{\partial\nu_{i}}=\frac{\mathsf{f}}{t_{z}+z}\frac{\partial x}{\partial\nu_{i}}-\frac{\mathsf{f}x}{\left(t_{z}+z\right)^{2}}\frac{\partial z}{\partial\nu_{i}},
\]
\[
\frac{\partial v}{\partial\nu_{i}}=\frac{\mathsf{f}}{t_{z}+z}\frac{\partial y}{\partial\nu_{i}}-\frac{\mathsf{f}y}{\left(t_{z}+z\right)^{2}}\frac{\partial z}{\partial\nu_{i}}.
\]
These are the velocities of the vertices of the quadrilaterals
\[
\frac{\partial Q}{\partial\nu_{i}}\equiv\left\{ \left(\frac{\partial u_{1}}{\partial\nu_{i}},\frac{\partial v_{1}}{\partial\nu_{i}}\right),\dots\right\}
\]
that define the boundaries of the integration ($\text{\ref{eq:pixel}}$).
The rate of change of that integral can be expressed as a line integral over the boundary
\begin{equation}
\frac{\partial \bar{I}\left(u,v,Q\right)}{\partial\nu_{i}}=I\left(Q\right)\underset{\partial Q}{\oint}V_{n}\left(l\right)K\left(u,v,u'\left(l\right),v'\left(l\right)\right)dl,\label{eq:dpixel}
\end{equation}
where $V_{n}$ is the velocity of a point on the boundary projected onto the boundary's outward normal. This integration can be performed as follows. Each edge (for example, between points 1 and 2) is parameterized by $t\in\left[0,1\right]$
\begin{equation}
\begin{cases}
u'\left(t\right)=\left(1-t\right)u_{1}+tu_{2}\\
v'\left(t\right)=\left(1-t\right)v_{1}+tv_{2}
\end{cases}\label{eq:u,v}
\end{equation}
The velocity of points on this edge is a linear combination
of velocities of its vertices
\[
\vec{V}\left(t\right)=\left(1-t\right)\left(\frac{\partial u_{1}}{\partial\nu_{i}},\frac{\partial v_{1}}{\partial\nu_{i}}\right)+t\left(\frac{\partial u_{2}}{\partial\nu_{i}},\frac{\partial v_{2}}{\partial\nu_{i}}\right),
\]
and the normal vector is
\[
\vec{n}=\frac{\left(v_{1}-v_{2},u_{2}-u_{1}\right)}{\sqrt{(v_{1}-v_{2})^{2}+(u_{2}-u_{1})^{2}}}.
\]
Thus, the contribution to the integral ($\text{\ref{eq:dpixel}}$) from this edge is
\begin{equation}
\int_{0}^{1}\left(\vec{V}\left(t\right),\vec{n}\right)K\left(u,v,u'\left(t\right),v'\left(t\right)\right)\frac{dl}{dt}dt,\label{eq:dpixel_edge}
\end{equation}
where
\[
\frac{dl}{dt}=\sqrt{(u_{2}-u_{1})^{2}+(v_{2}-v_{1})^{2}}.
\]
The arguments of the Gaussian kernel $K\left(u,v,u'\left(t\right),v'\left(t\right)\right)$
and $\left(\vec{V}\left(t\right),\vec{n}\right)$ are linear functions of $t$. This allows the integral ($\text{\ref{eq:dpixel_edge}}$)
to be evaluated symbolically in terms of the error function
\[
\textrm{erf }x=\frac{2}{\sqrt{\pi}}\int_{0}^{x}e^{-t^{2}}dt.
\]
By combining contributions from all edges of each visible quadrilateral, we obtain
\[
\frac{\partial \bar{I}\left(u,v\right)}{\partial\nu_{i}}=\sum_{\text{visible }Q}\frac{\partial \bar{I}\left(u,v,Q\right)}{\partial\nu_{i}}.
\]
The derivative of the image is
\[
\frac{\partial\mathcal{I}}{\partial\nu_{i}}=\left\{ \frac{\partial \bar{I}\left(u,v\right)}{\partial\nu_{i}}\middle|\, u,v\in\left\{ -1,-0.95,\dots,1\right\} \right\} .
\]
\end{document}